%% file: checklist.tex
\documentclass{article}

\usepackage[table]{xcolor}

\usepackage[preprint]{neurips_2026}

\usepackage[utf8]{inputenc}
\usepackage[T1]{fontenc}
\usepackage{hyperref}
\usepackage{url}
\usepackage{booktabs}
\usepackage{amsmath}
\usepackage{amsfonts}
\usepackage{nicefrac}
\usepackage{microtype}
\usepackage{xcolor}
\usepackage{pifont}
\usepackage{multirow}
\usepackage{array}
\usepackage{float}
\usepackage{wrapfig}
\usepackage{enumitem}
\usepackage{listings}
\usepackage{amssymb}
\usepackage{graphicx}
\usepackage{titletoc}

\hypersetup{
  colorlinks=true,
  linkcolor=red,
  citecolor=blue,
  urlcolor=magenta
}
\newcommand{\cmark}{\textcolor{green!60!black}{\ding{51}}}
\newcommand{\xmark}{\textcolor{red!70!black}{\ding{55}}}
\newcommand{\bench}{\textsc{DDx-TRACE}}
\newcommand{\up}[1]{\,\textcolor{green!60!black}{\ensuremath{\blacktriangle}\,#1}}
\newcommand{\down}[1]{\,\textcolor{red!70!black}{\ensuremath{\blacktriangledown}\,#1}}

\definecolor{jsonkey}{rgb}{0,0.5,0.5}
\definecolor{jsonvalue}{rgb}{0.6,0,0}

\lstdefinelanguage{json}{
    basicstyle=\small\ttfamily,
    showstringspaces=false,
    breaklines=true,
    upquote=true,
    morestring=[b]",
    stringstyle=\color{jsonvalue},
    literate=
     *{0}{{{0}}}{1}
      {1}{{{1}}}{1}
      {2}{{{2}}}{1}
      {3}{{{3}}}{1}
      {4}{{{4}}}{1}
      {5}{{{5}}}{1}
      {6}{{{6}}}{1}
      {7}{{{7}}}{1}
      {8}{{{8}}}{1}
      {9}{{{9}}}{1}
      {:}{{{\color{black}{:}}}}{1}
      {,}{{{\color{black}{,}}}}{1}
      {\{}{{{\color{black}{\{}}}}{1}
      {\}}{{{\color{black}{\}}}}}{1}
      {[}{{{\color{black}{[}}}}{1}
      {]}{{{\color{black}{]}}}}{1},
}

\title{\bench{}: A Benchmark for Medical Diagnostic Trajectories in VLMs}

\author{%
\textbf{Jiazhen Pan}$^{1,2,3,*}$ \quad
\textbf{Weixiang Shen}$^{1,2,4,*}$ \quad
\textbf{Jun Li}$^{1,3,\dagger}$ \quad
\textbf{Julian Canisius}$^{2}$ \\
\textbf{Felix Bitzer}$^{2}$ \quad
\textbf{Paula Roßmüller}$^{2}$ \quad
\textbf{Jiancheng Yang}$^{5}$ \quad
\textbf{Virginie Kreutzinger}$^{2}$ \\
\textbf{Daniel Rueckert}$^{1,2,3,6}$ \quad
\textbf{Benedikt Wiestler}$^{2,3}$
\\\\
$^1$Technical University of Munich (TUM)\quad
$^2$TUM University Hospital \\
$^3$Munich Center for Machine Learning (MCML) \quad
$^4$LMU Munich \\
$^5$Aalto University \quad
$^6$Imperial College London\\[0.2em]
$^{*}$Equal contribution  \quad $^{\dagger}$Corresponding author\\
}

\begin{document}

\maketitle

\begin{center}
\textbf{Code:} \url{https://github.com/JakobShen/DDx-TRACE}\\
\textbf{Dataset:} \url{https://huggingface.co/datasets/User3033/DDx-TRACE}
\end{center}

\begin{abstract}

Medical diagnosis is not a single prediction from a fully specified vignette. It is a sequential workup: clinicians decide what evidence to obtain, revise a differential diagnosis, and stop when the diagnosis is sufficiently supported. Most medical AI benchmarks instead reveal the relevant context upfront and score only the final answer, making unsupported correct guesses, premature closure, inefficient workups, and poor uncertainty updating invisible.
We introduce \bench{}, a physician-adjudicated benchmark for multimodal neuroradiology that evaluates diagnostic trajectories under hidden evidence over 211 challenging cases. Each case begins with limited clinical history; models request imaging studies in free form, receive matched image bundles when available, update a probabilistic differential diagnosis after each turn, and stop with a localized final diagnosis.
Evaluating state-of-the-art VLMs, we find that final diagnosis scores can substantially misrepresent workup quality: models may guess plausible diagnoses without essential evidence, request useful studies but misinterpret raw images, or acquire evidence inefficiently while updating uncertainty poorly. Controlled evidence variants isolate bottlenecks in planning, visual evidence extraction, and downstream differential reasoning. \bench{} shifts medical AI evaluation from final answers to evidence-supported diagnostic trajectories.

\end{abstract}

\section{Introduction}

\begin{figure*}[t!]
    \centering
    \includegraphics[width=\linewidth]{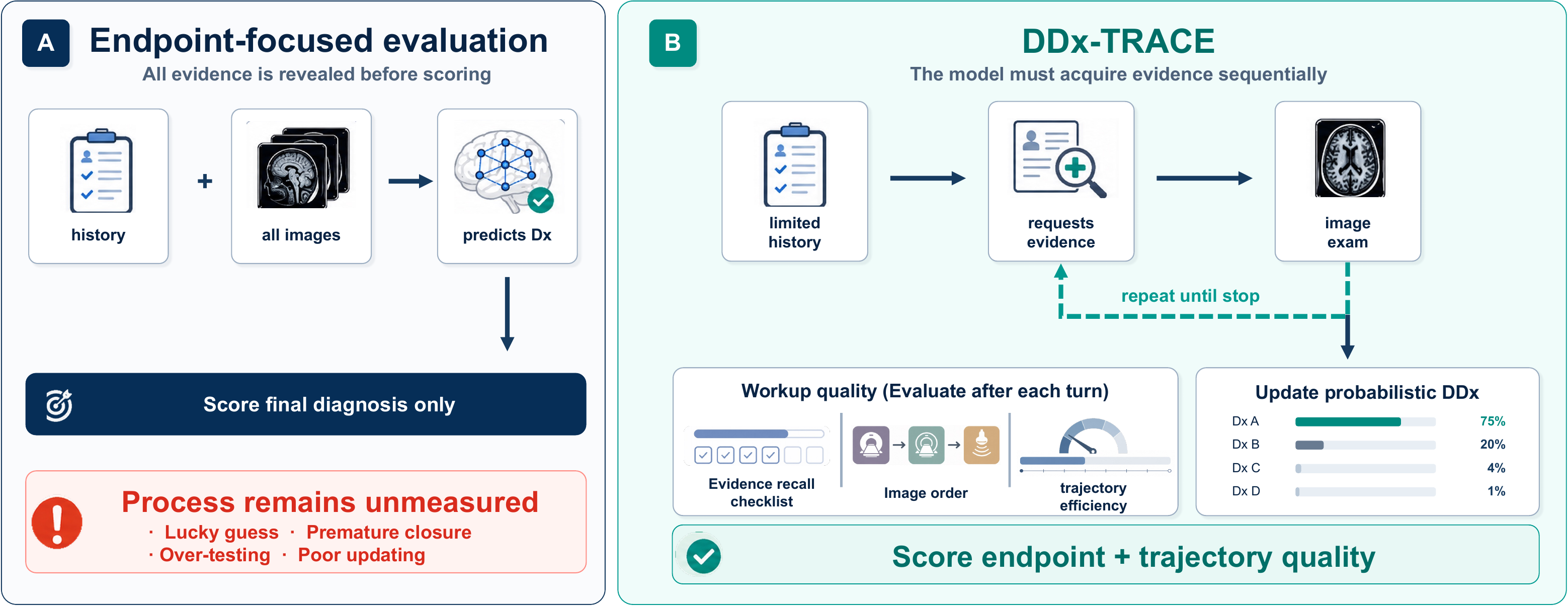}
    \caption{\textbf{\bench{} overview.} A) Conventional medical benchmarks often reveal \textit{all} the relevant evidence upfront and score only the final answer, making it difficult to detect unsupported correct guesses, premature closure, over-testing, or poor belief updating. B) \bench{} instead starts from a \textit{limited} history and requires the model to request imaging evidence sequentially, update a probabilistic differential diagnosis after each turn, and stop with a localized diagnosis. This makes the diagnostic trajectory itself measurable through endpoint, route, efficiency, and confidence metrics. A detailed pictorial illustration of \bench{} is presented in Appendix Fig. \ref{fig:workflow}.}
    \label{fig:overview}
\end{figure*}

Large language models (LLMs) and vision-language models (VLMs) have become increasingly competent at medical question answering \cite{singhal2023large,singhal2025toward,zhao2026agentic}, image interpretation \cite{li2023llava,chen2024vision}, report generation \cite{hartsock2024vision,saab2024capabilities,tanno2025collaboration}, and diagnostic reasoning tasks \cite{zhang2023huatuogpt,chen2024huatuogpt,pan2025medvlm}. Currently, most benchmarks for these models assess whether they can produce the correct answer when given a fixed input: a clinical vignette, an image, a report, or a set of retrieved findings. This endpoint-oriented evaluation, however, ignores a central part of clinical diagnosis. In practice, diagnosis is not only a prediction problem over observed evidence; it is a sequential decision-making process in which clinicians decide what evidence to acquire, how to prioritize it, how to revise the list of differential diagnoses, and when the available evidence is sufficient to make a final diagnosis.

Existing benchmarks cover important pieces of medical reasoning, including fixed-evidence QA \cite{jiang2025medagentbench,fan2025ai}, image-conditioned diagnosis \cite{zuo2025medxpertqa,Yaoetal2026}, and text-based clinical interaction \cite{hager2024evaluation,chiu2025simulating}. However, these settings still largely evaluate diagnosis after the evidentiary context has been defined for the model. The missing evaluation target is the diagnostic workup itself: a clinically grounded, partial-information workflow in which a model must decide what evidence to acquire, interpret ordered image scans, update a probabilistic differential diagnosis as evidence accumulates, and arrive at a localized final diagnosis through the route it followed. A clinically useful diagnostic model should not merely name a disease after seeing all available evidence. Rather, it should request relevant diagnostic studies, avoid low-value or unavailable requests, integrate findings across modalities and sequences, update uncertainty over time, and stop only when the workup is sufficiently supported. Endpoint accuracy alone cannot distinguish a clinically appropriate workup from a lucky guess. A model may produce the correct final diagnosis while missing physician-judged essential evidence. Conversely, it may request the right study but fail to interpret the ordered images. Such failures are largely invisible to benchmarks that reveal evidence upfront and score only the final answer.

We introduce \textbf{\bench{}}, a physician-annotated benchmark for evaluating \emph{differential-diagnosis trajectories} in multimodal neuroradiology. Each case begins with limited patient history only - like in clinical reality. The model does not receive an inventory of available studies, candidate diagnoses, expert findings, or the final label. Instead, at each turn, it issues a free-form imaging request, receives the matched image bundle if available, and updates a probabilistic differential diagnosis list. A case ends when the model stops and submits a localized final diagnosis. This protocol makes the diagnostic trajectory observable: what evidence the model seeks, what it ignores, how it updates its beliefs, and whether its final answer is supported by the acquired evidence.
\bench{} is constructed from 211 curated EuroRad-derived \cite{eurorad_database} neuroradiology cases containing 785 requestable imaging evidence units and 1,609 images. Physicians annotate each case with exam-level importance labels, preferred workup order, difficulty, rarity, corrected metadata, and case-specific diagnosis and localization rubrics. These annotations support process-aware evaluation beyond endpoint correctness. In addition to final diagnosis, localization, and differential-list quality, \bench{} measures essential-evidence recall, workup-order concordance, unmatched request rate, optional-evidence burden, stopping behavior, and confidence-weighted belief updating over the trajectory.

We use \bench{} to evaluate frontier, open-weight, and medical/radiology-adapted VLMs. Our results reveal substantial gaps between final diagnostic performance and clinically grounded diagnostic decision-making. Models with similar endpoint scores can differ substantially in whether they acquire essential evidence, follow physician-preferred study order, avoid unmatched requests, and update their differential diagnosis appropriately. Controlled evidence variants further decompose failures into planning, visual evidence extraction, and downstream differential reasoning: revealing all images, revealing studies in gold order, or providing oracle findings improves different models in different ways. These results suggest that current multimodal diagnostic agents remain limited not only by medical knowledge, but also by active evidence acquisition, image-to-finding extraction, and uncertainty-aware reasoning - barriers that are important to address before these models can be actively deployed in clinical decision-making. \bench{} therefore shifts medical AI evaluation from asking only \emph{what diagnosis did the model give?} to also asking \emph{how did it get there, and was the workup clinically sufficient?} Our contributions are threefold:
\begin{enumerate}[leftmargin=1.3em, topsep=1pt, itemsep=0pt, parsep=0pt, partopsep=0pt]
    \item \textbf{Task formulation.} We reformulate multimodal diagnosis as a hidden-evidence, turn-based diagnostic workup in which models request imaging evidence from limited history, update a probabilistic differential diagnosis after each turn, and stop with a localized final diagnosis.

    \item \textbf{Benchmark.} We introduce \bench{}, a physician-adjudicated neuroradiology benchmark built from curated EuroRad-derived cases, with requestable imaging bundles, evidence-importance labels, preferred workup order, and case-specific diagnosis and localization rubrics.

    \item \textbf{Evaluation and findings.} We define process-aware metrics for endpoint quality, essential-evidence recall, workup order, request efficiency, stopping behavior, and confidence alignment, and use them to show that final-answer performance can substantially misrepresent diagnostic workup quality in current multimodal models.

\end{enumerate}

\section{Related Work and Positioning}
\label{sec:related_work}

\textbf{Early medical LLMs/VLMs benchmarks} largely frame clinical reasoning as a fixed-input prediction problem. Text-based benchmarks such as MedQA, PubMedQA, MedMCQA, and medical subsets of general knowledge exams evaluate whether a model can answer medical questions or select the correct option from a static prompt \cite{jin2021disease,jin2019pubmedqa,pal2022medmcqa,qiu2024towards}. Multimodal benchmarks such as VQA-RAD \cite{lau2018dataset}, PathVQA \cite{he2020pathvqa}, SLAKE \cite{liu2021slake}, MedXpertQA \cite{zuo2025medxpertqa} and related medical VQA datasets \cite{lin2023medical} extend this setting to images, but still typically provide the visual evidence upfront and score the final answer, report, or classification output.
These benchmarks have been essential for measuring medical knowledge, image recognition, and fixed-evidence reasoning, but they do not directly evaluate whether a model can decide what clinical evidence to acquire, when to acquire it, or when the workup is sufficient to support a diagnosis.

\begin{table*}
\centering
\footnotesize
\setlength{\tabcolsep}{2pt}
\renewcommand{\arraystretch}{1.10}
\begin{tabular}{@{}p{0.18\linewidth}>
{\centering\arraybackslash}p{0.07\linewidth}>{\centering\arraybackslash}p{0.10\linewidth}>{\centering\arraybackslash}p{0.09\linewidth}>{\centering\arraybackslash}p{0.08\linewidth}>{\centering\arraybackslash}p{0.08\linewidth}>{\centering\arraybackslash}p{0.10\linewidth}p{0.25\linewidth}@{}}
\toprule
\shortstack{\textbf{Benchmark}\\\textbf{beyond simple QA}} & \shortstack{\textbf{Multi}\\\textbf{modal}} & \shortstack{\textbf{Open-ended}\\\textbf{acquisition}} & \shortstack{\textbf{Step}\\\textbf{importance}} & \shortstack{\textbf{Exam}\\\textbf{order}} & \shortstack{\textbf{Route}\\\textbf{Efficiency}} & \shortstack{\textbf{DDx}\\\textbf{confidence}} & \shortstack{\textbf{Main gap relative to} \\ \textbf{\bench{}}} \\
\midrule
Hager et al.~\cite{hager2024evaluation} & \xmark & \cmark & \xmark & Limited & \xmark & \xmark & Text only interaction; No route labels; endpoint-focused   \\
MedHELM~\cite{bedi2026holistic} & \xmark & \xmark & \xmark & \xmark & \xmark & Limited &  Fixed prompt; single turn; endpoint-focused \\
VivaBench~\cite{chiu2025simulating} & \xmark & \cmark & \xmark &  \xmark & Limited & \cmark & Text only interaction; limited route labels; endpoint-focused \\
MedThinkVQA~\cite{Yaoetal2026} & \cmark & \xmark & \xmark & \xmark & Limited & Limited & No active evidence acquisition; endpoint-focused\\
Healthbench~\cite{arora2025healthbench} & \xmark & Limited & \cmark & Limited & \xmark & \xmark & Text only interaction; fixed prompt; \\
\midrule
\textbf{\bench{} (ours)} & \cmark & \cmark & \cmark & \cmark & \cmark & \cmark & \\
\bottomrule
\end{tabular}
\caption{\textbf{Positioning of \bench{} among representative medical benchmarks.}
Prior benchmarks evaluate important subsets of medical reasoning, such as fixed-input QA, text-based interaction, or multi-image reasoning. However, they do not evaluate in a clinically grounded diagnostic-workup setting, e.g. with hidden raw imaging evidence, open-ended evidence acquisition, physician-labeled workup route quality, differential diagnosis updates, and process-aware trajectory metrics.}
\label{tab:benchmark_comparison}
\end{table*}

More recent benchmarks move \textbf{beyond simple endpoint QA} by testing broader clinical reasoning \cite{khandekar2024medcalc,wu2025medcasereasoning,bedi2026holistic}, medical conversational interaction \cite{hager2024evaluation,arora2025healthbench,zhu2025ask}, stress testing in medical context \cite{chang2025red,pan2025beyond}, multi-step diagnosis \cite{chiu2025simulating,nori2025sequential}, or multi-image interpretation \cite{chen2024towards,Yaoetal2026}.
These works motivate the need to evaluate models under more realistic clinical constraints, but \emph{existing benchmarks typically cover only part of the diagnostic workup loop}: some emphasize text-based information gathering without raw imaging, some evaluate multimodal reasoning after the evidence is already provided, and others score clinical responses without physician-labeled step importance or exam order. In contrast, \bench{} evaluates the full imaging-driven diagnostic trajectory: the model starts from limited history, requests imaging evidence in an open-ended setting, updates a differential diagnosis after each turn, and is scored on final diagnosis/localization as well as essential-evidence recall, workup order, efficiency, and confidence alignment. Table~\ref{tab:benchmark_comparison} summarizes this positioning.

\section{Benchmark Construction}
\label{sec:construction}

\paragraph{Source cases and scope.} \bench{} is developed upon EuroRad \cite{eurorad_database} cases published under a CC BY-NC-SA 4.0 license. We focus on neuroradiology because it naturally stresses three central abilities: deciding what imaging evidence to request next, integrating information across multiple modalities or sequences, and producing a localized final diagnosis rather than only a disease name: many cases cannot be reduced to recognizing a single image pattern. The diagnostic value lies in the sequence of studies and the evolution of the differential diagnosis reasoning process.

Starting from EuroRad teaching cases, we retain cases that include sufficient clinical history, at least 3 individual imaging exam bundles, and sufficient diagnostic context to support sequential workup annotation and rubric-based scoring. The current benchmark contains 211 curated cases, comprising 785 imaging exam bundles and 1,609 images across 28 modalities. The median case contains 4 imaging exam bundles and 8 images, reflecting the fact that many cases require aggregating evidence across multiple studies rather than solving the task from a single decisive image. Cases flagged as follow-up-only or post-treatment-only are discarded in this work. A more detailed benchmark statistic is documented in Appendix \ref{sec:benchmark_statistics}.

We reinterpret each retained case as a hidden inventory of requestable imaging exam bundles. Concretely, each figure or figure group is treated as an imaging exam bundle, the individual atomic unit that a model can request, and the benchmark asks whether a model can reconstruct a clinically supported workup from limited initial information e.g. patient history. This design preserves the educational and diagnostic richness of the source material while converting a published case report into a turn-based evaluation problem.

\paragraph{Case schema.} Each final case contains five components: (i) patient background and clinical history, the only information shown to the model at turn \(t=0\); (ii) imaging exam bundles, including image paths and metadata such as modality, acquisition, view, region, and contrast; (iii) an expert differential diagnosis list, used only for evaluation; (iv) the final diagnosis; and (v) case-specific diagnosis and localization rubrics. This structure supports both endpoint scoring, via the final diagnosis, differential list, and rubrics, and route scoring, via the requested imaging bundles and physician annotations of evidence importance and recommended workup sequence. The code is released at \url{https://github.com/JakobShen/DDx-TRACE}, and the dataset is released at \url{https://huggingface.co/datasets/User3033/DDx-TRACE}

\subsection{Clinician annotation protocol}
To standardize heterogeneous EuroRad cases, we first used an LLM (GPT-5.2) to draft a structured template for each case. This draft template proposed a normalized schema for patient history, imaging exams, figure-level evidence units, differential diagnoses, and endpoint rubrics. The LLM output was used only as a drafting aid; all clinically consequential fields were reviewed and corrected by physicians before entering the final benchmark release. Five board-certified physicians were involved in the annotation process.

In each case, at least two physicians independently annotated five aspects. \textbf{(1)}, they labeled the importance of intermediate imaging steps using the final label set \{\texttt{essential}, \texttt{optional}, \texttt{unnecessary}\}. \textbf{(2)}, they assigned a preferred order over imaging exams, allowing ties when two studies occupy the same clinically supported stage of the workup. \textbf{(3)}, they labeled case rarity and case difficulty. \textbf{(4)}, they corrected exam-level metadata, including modality, acquisition, view, imaged region, temporal context, and contrast usage. \textbf{(5)}, they corrected template artifacts such as modality naming errors, weak rubric criteria, or mismatches between figures and structured fields.

In the current release, 211 cases received at least two physician annotations and a meta-review pass, yielding 430 physician annotations in total. Agreement details and adjudication statistics are detailed in the Appendix~\ref{app:annotation_quality}. Generally, disagreement was concentrated in borderline distinctions such as \texttt{essential} versus \texttt{optional}, adjacent-rank swaps in preferred order, and \texttt{rare} versus \texttt{extreme rare}. These disagreements were escalated to board-certified senior physicians, who made the final labeling decision. We also find that LLM template artifacts were common enough to require active clinician correction, reinforcing the need to treat the benchmark schema as physician-authored rather than automatically extracted. In the present evaluation, we selected 191 of the 211 cases for further experiments and excluded the 20 cases that were contradictory and led to disagreement even during the senior physician's review; these cases, however, are still included in the released dataset.

\section{Task Formulation and Evaluation Protocol}
\label{sec:protocol}

\paragraph{Sequential evidence acquisition.}
\bench{} is formulated as a partial-information, sequential evidence-acquisition task (Fig.~\ref{fig:overview} and Fig. \ref{fig:workflow}). At turn \(t=0\), the model receives only patient background and clinical history. The hidden exam inventory, candidate diagnoses, expert imaging findings, final diagnosis, and gold labels are not revealed. At each turn, the model may either request one imaging examination in free-form natural language or stop. A request resolver returns at most one matched imaging exam bundle with minimal metadata; unmatched requests return no evidence but remain in the trajectory. The scored evidence unit is the imaging exam bundle rather than an individual image slice. After each request resolution, the model outputs an updated four-item differential diagnosis with probabilities summing to 1. On the stop turn, the same list is treated as the final differential diagnosis, and the model additionally provides a structured localization answer. This open-ended protocol tests whether a model can decide what evidence to acquire, not only whether it can recognize abnormalities once images are shown.

\paragraph{Outputs and gold route labels.}
The official output schema contains \texttt{current\_differential}, \texttt{action}, \texttt{requested\_examination} when applicable, and \texttt{final\_location} on the stop turn. Gold annotations map figure-linked diagnostic steps to requestable imaging exam bundles with importance and preferred-order labels. Exam bundles without explicit diagnostic-step annotations are treated as optional in route metrics. Because some cases contain tied preferred-order stages, the gold workup is represented as a partial order rather than a strict total ranking. Follow-up, postoperative, and post-treatment studies are excluded from the official request pool and route metrics. Further schema and matching details are provided in Appendix~\ref{app:protocol_details}.

\subsection{Metrics}
\label{sec:metrics}

We evaluate both the final diagnostic answer and the diagnostic trajectory. All main results are reported as decomposed metrics rather than a single official composite score, because models with similar endpoint scores can differ substantially in evidence recall, request validity, workup order, and confidence behavior.

\paragraph{Endpoint metrics.}
We define three endpoint scores:
\begin{equation}
S_{\mathrm{dx}} \in [0,1], \qquad
S_{\mathrm{loc}} \in [0,1], \qquad
S_{\mathrm{ddx}} \in [0,1].
\label{eq:endpoint_scores}
\end{equation}
Here \(S_{\mathrm{dx}}\) scores the final top-1 diagnosis, \(S_{\mathrm{loc}}\) scores the final localization, and \(S_{\mathrm{ddx}}\) scores the final four-item differential diagnosis as a clinically plausible ranked set. These targets are separated because a model may localize a lesion while missing the disease label, or name the correct disease while assigning implausible probability mass to the rest of the differential.

\paragraph{Route metrics.}
Let \(R_m\) be the set of matched requested exams, \(R_u\) the set of unmatched requests, \(G_{\mathrm{ess}}\) the set of gold essential exams, and \(G_{\mathrm{opt}}\) the set of optional exams, including unlabeled requestable bundles. We report:
\begin{equation}
S_{\mathrm{ER}} =
\frac{|R_m \cap G_{\mathrm{ess}}|}{|G_{\mathrm{ess}}|},
\qquad
B_{\mathrm{opt}} =
\frac{|R_m \cap G_{\mathrm{opt}}|}{\max(1, |R_m|)},
\qquad
B_{\mathrm{unm}} =
\frac{|R_u|}{\max(1, |R_m| + |R_u|)}.
\label{eq:route_basic}
\end{equation}
\(S_{\mathrm{ER}}\) measures essential-evidence recall, \(B_{\mathrm{opt}}\) measures optional burden, and \(B_{\mathrm{unm}}\) measures unmatched-request rate. Higher is better for \(S_{\mathrm{ER}}\), whereas lower is better for both burden metrics.

For route ordering, let \(r(e)\) be the gold order stage of exam \(e\), and let \(t(e)\) be the turn at which matched exam \(e\) is requested. We define the comparable precedence set
\[
\mathcal{P}_R =
\{(e_i,e_j) : e_i,e_j \in R_m \cap (G_{\mathrm{ess}} \cup G_{\mathrm{opt}}), \; r(e_i) < r(e_j)\}.
\]
The order concordance score is
\begin{equation}
S_{\mathrm{order}} =
\frac{1}{|\mathcal{P}_R|}
\sum_{(e_i,e_j)\in \mathcal{P}_R}
\mathbf{1}[t(e_i) < t(e_j)],
\label{eq:order}
\end{equation}
computed only for cases with at least one comparable pair. Tied gold stages impose no precedence constraint.

\paragraph{Efficiency and clinical sufficiency.}
We use an imaging-request budget of \(B=6\). Let \(S_{\mathrm{dx}}^{(t)}\) be the normalized diagnosis score of the model's top diagnosis at turn \(t\), and let \(\tau\) be the threshold for a sufficiently correct diagnostic guess; in our experiments, \(\tau=2/3\). We define:
\begin{align}
T_{\mathrm{guess}}
&=
\min \left\{
t : S_{\mathrm{dx}}^{(t)} \geq \tau
\right\},
\label{eq:time_to_guess}
\\
T_{\mathrm{clin}}
&=
\min \left\{
t : S_{\mathrm{dx}}^{(t)} \geq \tau
\;\;\mathrm{and}\;\;
S_{\mathrm{ER}}^{(t)} = 1
\right\},
\label{eq:time_to_clinical}
\end{align}
where \(S_{\mathrm{ER}}^{(t)}\) is essential recall after the first \(t\) turns. \(T_{\mathrm{guess}}\) measures when the model first says the right thing, whereas \(T_{\mathrm{clin}}\) requires that the diagnosis is also supported by all physician-labeled essential evidence. If the condition is never met within the budget, the case is recorded as unreached. We additionally report the reached fraction $R_{\mathrm{clin}}$, number of requested examinations, and stopping turn in the appendix efficiency analysis.

\paragraph{Trajectory and confidence metrics.}
Because the model outputs a four-item differential diagnosis after every turn, \bench{} can evaluate how beliefs evolve as evidence accumulates. A rubric-conditioned judge labels each unique diagnosis string appearing in the trajectory as exact-match (\(E\)), acceptable differential (\(A\)), or unmatched (\(U\)). Let \(p_i^{(t)}\) be the probability assigned to diagnosis \(i\) at turn \(t\). We define:
\begin{equation}
S_{\mathrm{conf}}^{(t)}
=
\sum_{i \in E_t} p_i^{(t)}
+
\sum_{i \in A_t} p_i^{(t)}
-
\sum_{i \in U_t} p_i^{(t)},
\qquad
S_{\mathrm{traj}} =
\frac{1}{T}\sum_{t=1}^{T} S_{\mathrm{conf}}^{(t)}.
\label{eq:traj_conf}
\end{equation}
Here \(E_t\), \(A_t\), and \(U_t\) are the diagnoses in the turn-\(t\) differential assigned to each category, and \(T\) is the final turn. We report the final-turn value \(S_{\mathrm{conf}}=S_{\mathrm{conf}}^{(T)}\) as confidence alignment, and \(S_{\mathrm{traj}}\) as its trajectory average. These metrics reward probability mass on the exact diagnosis or clinically acceptable alternatives and penalize persistent confidence in unmatched diagnoses.

\paragraph{Scoring and reporting.}
Route metrics are computed after request-to-exam matching against the hidden exam inventory. Endpoint scoring uses a fixed rubric-conditioned LLM-as-a-judge, Gemini 3 Flash, with structured JSON output. The judge also labels unique diagnosis strings as exact, acceptable, or unmatched; these labels are reused for \(S_{\mathrm{conf}}\) and \(S_{\mathrm{traj}}\). Additional scorer details, including agreement with rule-based methods and human experts, are provided in Appendix~\ref{app:judge_reliability}.

\section{Experiments and Results}
\label{sec:experiments}

We evaluate 13 VLMs on the official \bench{} split, spanning frontier general-purpose VLMs, open-weight Qwen3.5 and Gemma models \cite{qwen2026qwen35collection,qwen2026qwen35modelcard,gemma3_2025}, and medical or radiology-adapted VLMs including MedGemma and Lingshu \cite{sellergren2025medgemma,xu2025lingshu}. All models follow the same hidden-evidence protocol (Sec.~\ref{sec:protocol}). Unless stated otherwise, metrics are macro-averaged over cases. Our central question is not only whether a model eventually names the diagnosis, but whether it reaches that diagnosis through a clinically sufficient workup. We therefore report endpoint, route, trajectory, and confidence metrics jointly, without defining a single official composite score.

\begin{table*}[t]
    \centering
    \small
    \makebox[\textwidth][c]{%
    \begin{tabular}{lccccccccc}
        \toprule
        & \multicolumn{6}{c}{\textbf{Process}} & \multicolumn{3}{c}{\textbf{Endpoint}} \\
        \cmidrule(lr){2-7} \cmidrule(lr){8-10}
        Model & \(S_{\text{ER}}\) $\uparrow$ & \(S_{\text{order}}\) $\uparrow$ & \(S_{\text{traj}}\) $\uparrow$ & \(S_{\text{conf}}\) $\uparrow$ & Opt.\ burden $\downarrow$ & Unm.\ req. $\downarrow$ & \(S_{\text{dx}}\) $\uparrow$ & \(S_{\text{loc}}\) $\uparrow$ & \(S_{\text{ddx}}\) $\uparrow$ \\
        \midrule
        Average & 0.573 & 0.395 & 0.286 & -0.051 & 0.245 & 0.261 & 0.304 & 0.425 & 0.364 \\
        \midrule
        \multicolumn{10}{c}{\small\textbf{Frontier VLMs}} \\
        \midrule
        GPT-5.4 & \cellcolor{green!8}0.716 & \cellcolor{green!15}0.574 & \cellcolor{green!8}0.407 & \cellcolor{green!8}0.183 & 0.245 & 0.268 & \cellcolor{green!8}0.430 & \cellcolor{green!25}\textbf{0.621} & \cellcolor{green!8}0.507 \\
        GPT-5.4 Mini & \cellcolor{red!8}0.425 & \cellcolor{red!15}0.178 & 0.334 & 0.043 & \cellcolor{green!25}\textbf{0.175} & \cellcolor{green!25}\textbf{0.103} & 0.346 & 0.482 & 0.408 \\
        Gemini 3 Flash & \cellcolor{green!25}\textbf{0.767} & \cellcolor{green!25}\textbf{0.619} & \cellcolor{green!15}0.498 & \cellcolor{green!25}\textbf{0.531} & \cellcolor{red!15}0.273 & 0.228 & \cellcolor{green!15}0.541 & \cellcolor{green!15}0.586 & \cellcolor{green!15}0.611 \\
        Gemini 3.1 Pro & \cellcolor{green!15}0.722 & \cellcolor{green!8}0.574 & \cellcolor{green!25}\textbf{0.501} & \cellcolor{green!15}0.504 & 0.242 & \cellcolor{green!15}0.143 & \cellcolor{green!25}\textbf{0.546} & \cellcolor{green!8}0.578 & \cellcolor{green!25}\textbf{0.644} \\
        Claude Sonnet 4.6 & 0.714 & 0.549 & 0.396 & 0.122 & \cellcolor{red!8}0.260 & 0.233 & 0.423 & 0.519 & 0.474 \\
        \midrule
        \multicolumn{10}{c}{\small\textbf{Open-weight models}} \\
        \midrule
        Qwen3.5-35B-A3B & \cellcolor{red!15}0.426 & \cellcolor{red!8}0.194 & 0.276 & 0.017 & 0.247 & \cellcolor{green!8}0.150 & 0.289 & 0.472 & 0.367 \\
        Qwen3.5-27B & 0.598 & 0.315 & 0.285 & -0.068 & \cellcolor{green!15}0.228 & 0.231 & 0.302 & 0.439 & 0.346 \\
        Qwen3.5-9B & 0.526 & 0.324 & 0.191 & -0.266 & \cellcolor{red!25}0.298 & 0.280 & 0.205 & 0.365 & 0.279 \\
        Qwen3.5-4B & 0.541 & 0.347 & \cellcolor{red!8}0.152 & \cellcolor{red!8}-0.357 & 0.254 & \cellcolor{red!15}0.392 & \cellcolor{red!8}0.163 & 0.282 & \cellcolor{red!8}0.202 \\
        Gemma 3 27B & 0.508 & 0.411 & 0.190 & -0.235 & 0.259 & \cellcolor{red!8}0.298 & 0.192 & \cellcolor{red!8}0.272 & 0.253 \\
        \midrule
        \multicolumn{10}{c}{\small\textbf{Medical / radiology-adapted VLMs}} \\
        \midrule
        MedGemma 27B & 0.579 & 0.444 & 0.163 & -0.267 & 0.238 & \cellcolor{red!25}0.538 & 0.168 & \cellcolor{red!15}0.264 & 0.241 \\
        MedGemma 1.5 4B & \cellcolor{red!25}0.260 & \cellcolor{red!25}0.103 & \cellcolor{red!25}0.080 & \cellcolor{red!25}-0.631 & 0.245 & 0.285 & \cellcolor{red!25}0.080 & \cellcolor{red!25}0.187 & \cellcolor{red!25}0.113 \\
        Lingshu-32B & 0.598 & 0.513 & \cellcolor{red!15}0.148 & \cellcolor{red!15}-0.423 & \cellcolor{green!8}0.235 & 0.275 & \cellcolor{red!15}0.152 & 0.300 & \cellcolor{red!15}0.175 \\
        \bottomrule
    \end{tabular}}
    \caption{\textbf{Main benchmark results.} \bench{} evaluates both the diagnostic route and the final answer. Green and red shading mark the top and bottom three models per metric, respectively; darker shading indicates stronger deviation, and bold denotes the best score.}
    \label{tab:main_results}
\end{table*}

\paragraph{Finding 1: Endpoint accuracy and diagnostic quality select different models.}
Table~\ref{tab:main_results} shows that \bench{} is far from saturated. Frontier models define the current upper limit, but even the best systems remain well below a reliable diagnostic workup standard.
Importantly, \emph{the ranking induced by final diagnosis score is not the ranking induced by process quality}.
Gemini 3.1 Pro obtains the best diagnosis and differential scores, \(S_{\mathrm{dx}}=0.546\) and \(S_{\mathrm{ddx}}=0.644\), while GPT-5.4 obtains the strongest localization score, \(S_{\mathrm{loc}}=0.621\). Process metrics identify a different leader: Gemini 3 Flash achieves the highest essential-evidence recall, order concordance, and confidence alignment, with \(S_{\mathrm{ER}}=0.767\), \(S_{\mathrm{order}}=0.619\), and \(S_{\mathrm{conf}}=0.531\). Thus, even among frontier models, endpoint quality and workup quality do not identify the same winner.

\begin{wrapfigure}{R}{0.45\textwidth}
    \centering
    \includegraphics[width=0.97\linewidth]{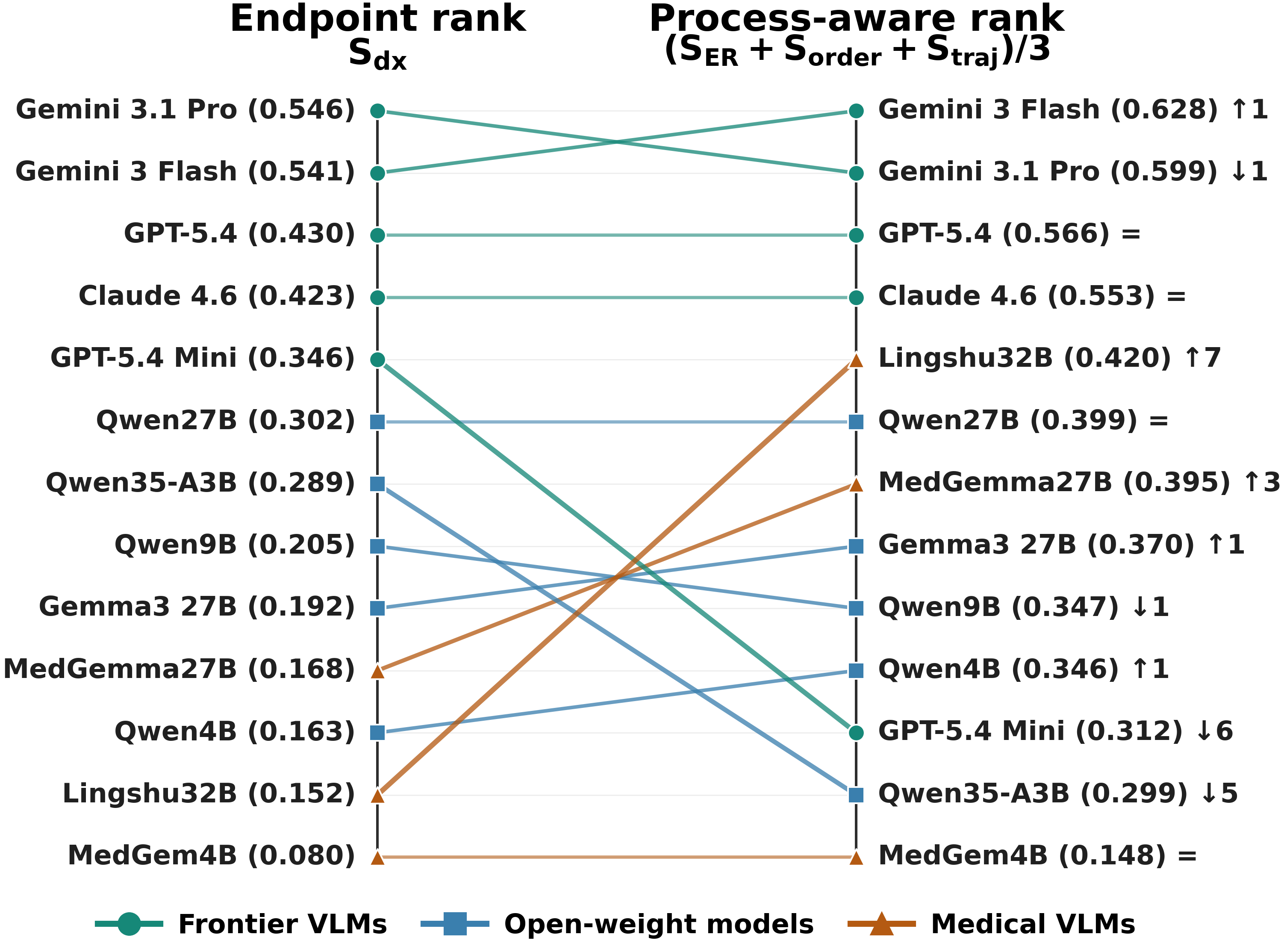}
    \caption{\textbf{Endpoint versus process-aware ranking.} Endpoint rank uses \(S_{\mathrm{dx}}\). Process rank is used only for visualization and is computed as the mean of \(S_{\mathrm{ER}}\), \(S_{\mathrm{order}}\), and \(S_{\mathrm{traj}}\).}
    \label{fig:ranking_change}
\end{wrapfigure}

This separation motivates route-aware evaluation. A correct final diagnosis may still be reached after missing essential evidence, requesting studies in a poor order, or stopping before the workup is sufficient. Conversely, useful evidence acquisition does not guarantee correct diagnostic integration. GPT-5.4 Mini, for example, has low unmatched-request and optional-burden rates, but weak essential recall and order concordance. MedGemma 27B and Lingshu-32B show the opposite pattern: they recover moderate essential evidence, yet remain weak on endpoint diagnosis and differential scores, suggesting failures in visual interpretation, evidence integration, or downstream reasoning.

Fig.~\ref{fig:ranking_change} makes this distinction explicit: GPT-5.4 Mini drops under process-aware ranking, whereas Lingshu and MedGemma 27B move upward despite weaker endpoint scores. The movement is moderate but important: endpoint scoring captures part of diagnostic behavior, but it cannot distinguish diagnostic guessers, premature stoppers, and models that request many studies without improving the final answer.

\paragraph{Finding 2: Passive evidence access hides active-acquisition and visual-extraction failures.}
We next isolate where failures arise, and compare the standard setting with controlled variants that remove one difficulty at a time: history-only removes imaging; all-images-at-once removes active acquisition; random-order and gold-order reveal remove model-chosen sequencing; and oracle findings add expert textual findings to matched image bundles. These variants are not alternative leaderboards, but probes for separating planning, evidence acquisition, visual extraction, and downstream reasoning.

\begin{table*}[t]
    \centering
    \small
    \setlength{\tabcolsep}{3pt}
    \begin{tabular}{lcccccc}
        \toprule
        Setting & \(S_{\text{dx}}\) & \(S_{\text{loc}}\) & \(S_{\text{ER}}\) & \(S_{\text{order}}\) & \(S_{\text{traj}}\) & Overall Change \\
        \midrule
        Default setting & 0.39 & 0.48 & 0.64 & 0.47 & 0.36 & -- \\
        \midrule
        History-only & 0.26\down{32.0} & 0.31\down{34.7} & -- & -- & 0.26\down{27.6} & \down{31.4} \\
        All-images-at-once & 0.42\up{9.3} & 0.50\up{5.4} & -- & -- & 0.42\up{17.1} & \up{10.6} \\
        Random-order reveal & 0.43\up{11.9} & 0.51\up{6.7} & -- & -- & 0.39\up{7.2} & \up{8.6} \\
        Gold-order reveal & 0.44\up{14.4} & 0.52\up{8.4} & -- & -- & 0.41\up{12.7} & \up{11.8} \\
        Oracle findings & 0.58\up{48.5} & 0.69\up{43.9} & 0.73\up{15.4} & 0.51\up{8.9} & 0.50\up{39.2} & \up{31.2} \\
        \bottomrule
    \end{tabular}
    \caption{\textbf{Bottleneck ablations.} Each row reports the mean score across representative models under a controlled evidence setting. Passive settings expose fixed evidence rather than model-requested exams, so route metrics are undefined and shown as ``--''. Overall Change is a descriptive mean relative change over available metrics; detailed per-model results are in Appendix~\ref{app:detailed_oracle_analysis}.}
    \label{tab:ablations_avg}
\end{table*}

\begin{table*}[th]
    \centering
    \scriptsize
    \setlength{\tabcolsep}{4pt}
    \begin{tabular}{lccccc}
        \toprule
        Model & \shortstack{\(T_{\mathrm{guess}}\) \\time to diagnostic \\ guess $\downarrow$} & \shortstack{\(R_{\mathrm{clin}}\)\\ Clinically supported\\ reached (\%) \(\uparrow\)} & \shortstack{\(T_{\mathrm{clin}}\mid\mathrm{reached}\)\\
Avg. acceptable time\\
successes only \(\downarrow\)} & \shortstack{Avg.\ exams\\requested} & \shortstack{Avg.\ turns\\until stopping} \\
        \midrule
        \multicolumn{6}{c}{\small\textit{Frontier VLMs}} \\
        \midrule
        GPT-5.4 & \cellcolor{green!8}5.437 & 15.8 & 3.600 & 3.421 & 4.421 \\
        GPT-5.4 Mini & 6.204 & 3.1 & 2.333 & 1.508 & 2.508 \\
        Gemini 3 Flash & \cellcolor{green!15}4.429 & \cellcolor{green!25}\textbf{30.9} & 3.695 & 3.675 & 4.675 \\
        Gemini 3.1 Pro & \cellcolor{green!25}\textbf{4.356} & \cellcolor{green!15}25.7 & 3.347 & 2.874 & 3.874 \\
        Claude Sonnet 4.6 & 5.658 & \cellcolor{green!8}16.3 & 3.742 & 3.405 & 4.405 \\
        \midrule
        \multicolumn{6}{c}{\small\textit{Open-weight models}} \\
        \midrule
        Qwen3.5-35B-A3B & 6.895 & \cellcolor{red!8}2.6 & 3.800 & 1.895 & 2.895 \\
        Qwen3.5-27B & 6.863 & 9.5 & 3.222 & 2.826 & 3.826 \\
        Qwen3.5-9B & 7.495 & \cellcolor{red!8}2.6 & 3.000 & 2.979 & 3.979 \\
        Qwen3.5-4B & \cellcolor{red!8}8.089 & 3.7 & 4.000 & 3.616 & 4.616 \\
        \midrule
        \multicolumn{6}{c}{\small\textit{Medical / radiology-adapted VLMs}} \\
        \midrule
        MedGemma 27B IT & 8.052 & \cellcolor{red!8}2.6 & 4.000 & 4.749 & 5.749 \\
        MedGemma 1.5 4B IT & \cellcolor{red!25}8.628 & \cellcolor{red!25}1.0 & 2.000 & 1.361 & 2.361 \\
        Lingshu-32B & \cellcolor{red!15}8.265 & \cellcolor{red!15}1.6 & 2.667 & 3.074 & 4.074 \\
        \bottomrule
    \end{tabular}
    \caption{
\textbf{Diagnostic efficiency and stopping behavior.}
\(T_{\mathrm{guess}}\) is the mean earliest turn at which the model's top diagnosis reaches the diagnosis-score threshold \(S_{\mathrm{dx}}^{(t)} \ge 2/3\). Unreached cases are assigned the sentinel value \(T_{\max}+1=9\). \(R_{\mathrm{clin}}\) is the clinically supported diagnosis reached rate: the fraction of cases in
which some turn \(t \le T_{\max}=8\) satisfies both \(S_{\mathrm{dx}}^{(t)} \ge 2/3\) and complete cumulative essential-evidence recall. \(T_{\mathrm{clin}}\mid\mathrm{reached}\) is the average first
acceptable turn among reached cases only; it excludes failures and should therefore be interpreted together with \(R_{\mathrm{clin}}\). Average exams requested and average stop turn describe stopping behavior and are not monotonic quality metrics.
}
    \label{tab:diagnostic_efficiency}
\end{table*}

Table~\ref{tab:ablations_avg} reveals a layered failure structure. First, removing active acquisition helps but does not solve the task: all-images-at-once raises diagnosis to 0.42, and gold-order reveal raises it to 0.44. Thus, fixed-evidence evaluation measures useful multimodal reasoning, but not whether the model can proactively plan and construct the evidence context needed for that reasoning. Second, Oracle findings produce the largest improvement. With expert textual findings attached to matched image bundles, diagnosis rises to 0.58, localization to 0.69, essential recall to 0.73, and trajectory score to 0.50. This points to \emph{visual evidence extraction as a major bottleneck}: models often fail not only by requesting the wrong study, but also by missing or underusing decisive findings in raw images. Yet oracle findings still do not close the benchmark, implying residual failures in differential reasoning, evidence prioritization, stopping, and confidence allocation. The ablations, therefore, show not just that current VLMs are inaccurate, but where the workup pipeline breaks.

\paragraph{Finding 3: Correct diagnostic guesses are rarely supported by complete essential evidence.}
Table~\ref{tab:diagnostic_efficiency} highlights that \emph{even the best model reaches the benchmark-defined clinically sufficient criterion in only 30.9\% of cases}. \(T_{\mathrm{guess}}\) is the first turn at which the model's top diagnosis reaches the diagnosis-score threshold \(\tau=2/3\), i.e., when the model first makes a sufficiently correct diagnostic guess. The clinically supported diagnosis reached rate is stricter: by \(T_{\max}=8\), the model must both meet this diagnostic threshold and acquire all physician-labeled essential evidence. The reported average acceptable time,
\(T_{\mathrm{clin}}\mid\mathrm{reached}\), is conditional on success; failures are excluded, so a low value is meaningful only when accompanied by a high reached rate.

The reached rate is strikingly low. Gemini 3 Flash is best but reaches a clinically supported diagnosis in only 30.9\% of cases; Gemini 3.1 Pro reaches 25.7\%, while GPT-5.4 and Claude Sonnet 4.6 remain near 16\%. Most systems are much lower: 8 of the 12 models in Table~\ref{tab:diagnostic_efficiency} have single-digit reached rates, including all medical/radiology-adapted VLMs. Thus, endpoint scoring can reward plausible or correct guesses long before the essential evidence needed to justify them has been acquired.

The conditional timing columns show why apparent efficiency should be interpreted cautiously. GPT-5.4 Mini and MedGemma 1.5 4B have short average acceptable times among reached cases, but reach clinical sufficiency in only 3.1\% and 1.0\% of cases, respectively; these are rare, easy successes, not reliable workups. Conversely, MedGemma 27B requests many exams and stops late, yet reaches clinical sufficiency in only 2.6\% of cases. \bench{} exposes this benchmark gap: current models can guess early, but seldom combine evidence acquisition, visual interpretation, belief updating, and stopping into a supported diagnostic trajectory. More studies of the efficiency-accuracy tradeoff, including the Pareto-frontier plot, can be found in Appendix \ref{sec:eff-acc}.

Fig.~\ref{fig:qualitative} illustrates this mismatch. In the caudal regression syndrome case, GPT-5.4 mini receives full endpoint credit for the top diagnosis, yet the route remains clinically insufficient: it makes an invalid request, misses the clinically relevant MRI examinations, and never reaches \(T_{\mathrm{clin}}\). The final answer is correct, but the workup is incomplete. This is the purpose of \bench{}: a route-aware evaluator should not treat a correct but unsupported answer as equivalent to a clinically grounded workup.

\begin{figure*}
    \centering
    \includegraphics[width=1\linewidth]{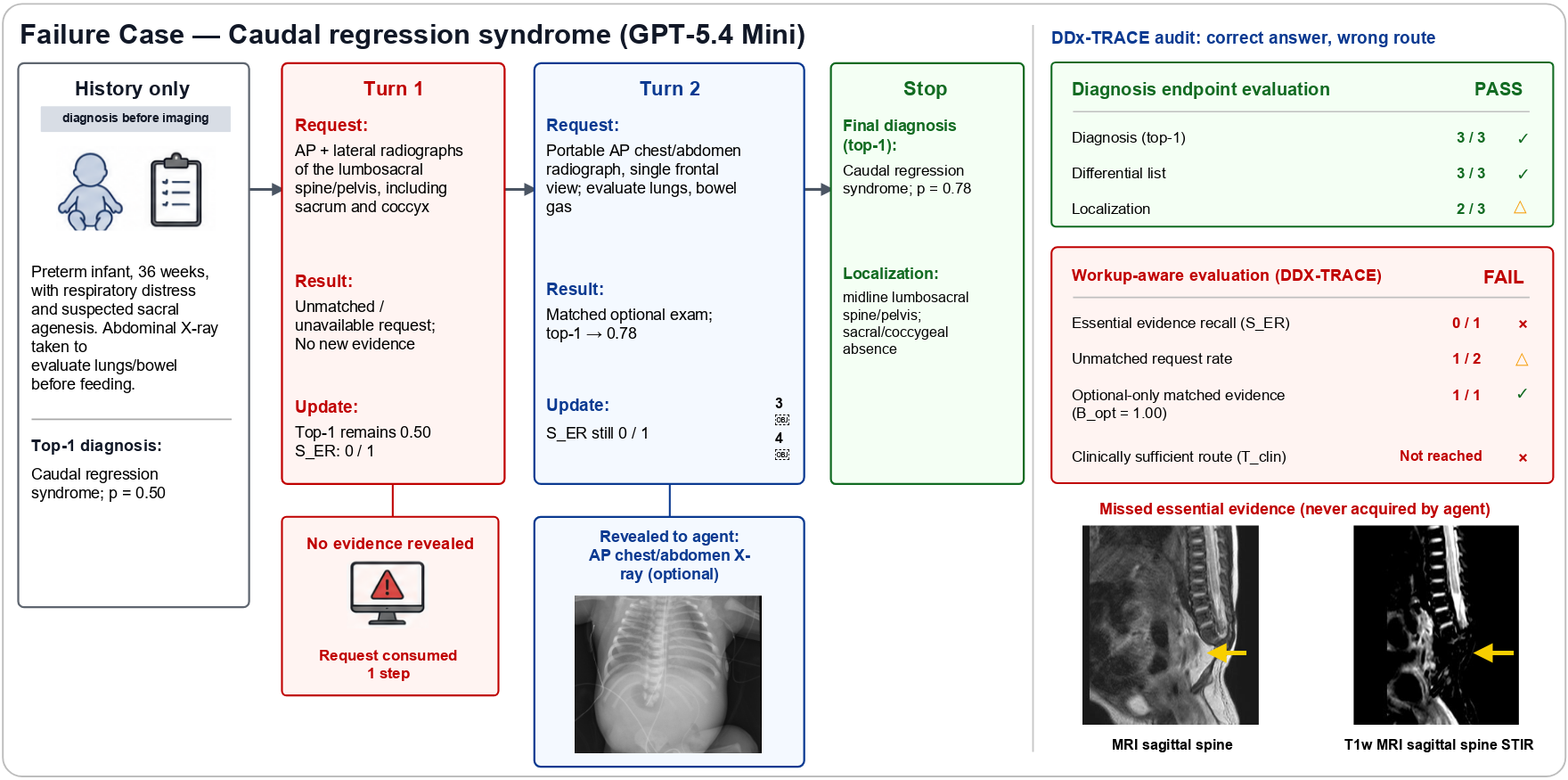}
    \caption{\textbf{Endpoint-pass/workup-fail audit.} A case-level trace contrasts final diagnostic credit with evidence acquisition, ordering, and clinical-sufficiency checks, showing how a correct answer can still arise from an insufficient diagnostic route.}
    \label{fig:qualitative}
\end{figure*}

In summary, current VLMs can sometimes guess the right diagnosis and sometimes request useful evidence, but they do not yet reliably combine evidence acquisition, visual interpretation, differential reasoning, and stopping into a clinically sufficient workup.

\section{Conclusion}
\label{sec:conclusion}

\bench{} evaluates multimodal diagnostic models as evidence-acquisition workups under partial information, rather than as endpoint answers to fully specified inputs. It changes the evaluation question from ``Can the model name the diagnosis after seeing all evidence?'' to ``Can the model decide what evidence to obtain, update uncertainty, and stop with a diagnosis supported by the workup?'' \bench{} is a controlled benchmark, not a deployment evaluation: it focuses on radiology cases with finite requestable imaging bundles, so route metrics should be interpreted within this benchmark environment rather than as direct measures of real-world clinical fitness or patient benefit. Its value is to expose whether VLM models can acquire/use evidence under controlled partial information. Our results show that current VLMs are substantially better at naming plausible diagnoses than completing supported diagnostic workups: correct or plausible answers often occur without complete physician-labeled essential evidence, a gap invisible to endpoint-focused benchmarks.

\section{Acknowledgement}
We would like to thank Kelly He for her support during the course of this project. This work is partially funded by the European Research Council (ERC) project Deep4MI (884622).

\clearpage
\bibliographystyle{plain}
\bibliography{refs}

\clearpage
\appendix

\include{appendix}

\clearpage
\end{document}

%% file: appendix.tex

\appendix

\section*{Appendix}
\startcontents[sections]
\printcontents[sections]{l}{1}{\setcounter{tocdepth}{2}}

\newpage
\section{Pictorial Illustration of \bench{} Workflow}

\begin{figure*}[th]
    \centering
    \includegraphics[width=\linewidth]{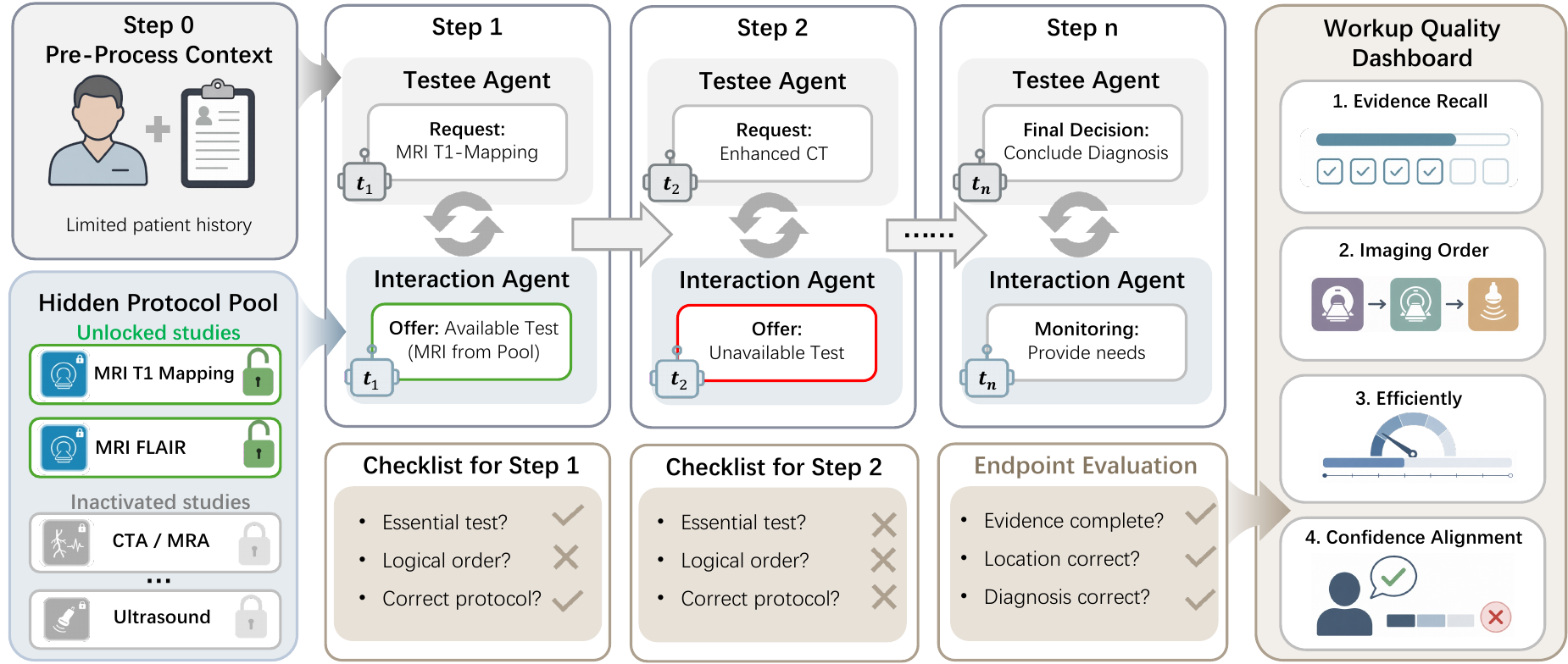}
    \caption{\textbf{\bench{} overview.} Each case begins with limited clinical history only, and the agent does not receive a list of available examinations. The agent interacts with the environment by requesting imaging exams in free form, observing newly revealed image bundles, updating a differential diagnosis list with probabilities after every turn, and finally producing a localized diagnosis. The benchmark evaluates endpoint correctness, route quality, and belief updating.}
    \label{fig:workflow}

\end{figure*}

\section{Limitations}
\label{app:limitations}

\bench{} is designed as a high-density evaluation benchmark for diagnostic-workup behavior rather than a training-scale dataset or a claim of clinical deployment readiness. Its current release focuses on neuroradiology, which provides a strong testbed for localization, modality/sequence selection, and multi-exam evidence integration, but also limits direct generalization to other radiology subspecialties, more common disease distributions, and non-radiology diagnostic workflows. Future extensions should broaden the case mix across organs, specialties, and care settings while preserving the physician-adjudicated trajectory annotations that make the benchmark process-aware. \bench{} also inherits the abstraction of EuroRad teaching cases: cases are retrospective, educationally curated, and represented by published key images rather than complete PACS/DICOM studies, full reports, laboratory data, physical examination findings, or longitudinal management information. This design makes controlled multimodal evaluation feasible, but future versions should incorporate fuller imaging studies and prospectively collected or institutionally diverse cases where licensing and privacy permit. Finally, the hidden-inventory protocol abstracts clinical ordering through a finite set of requestable imaging bundles. A model may therefore request a clinically reasonable study that is unavailable in the source case; such requests are counted as unmatched by the benchmark rather than necessarily clinically inappropriate. Route metrics should thus be interpreted as controlled signals of evidence acquisition, order, and efficiency within the benchmark environment, not as direct measures of real-world cost, guideline adherence, or patient benefit.

\newpage
\section{Benchmark Statistics and Data Distribution}
\label{sec:benchmark_statistics}

\begin{figure}[!htbp]
    \centering
    \includegraphics[width=1\linewidth]{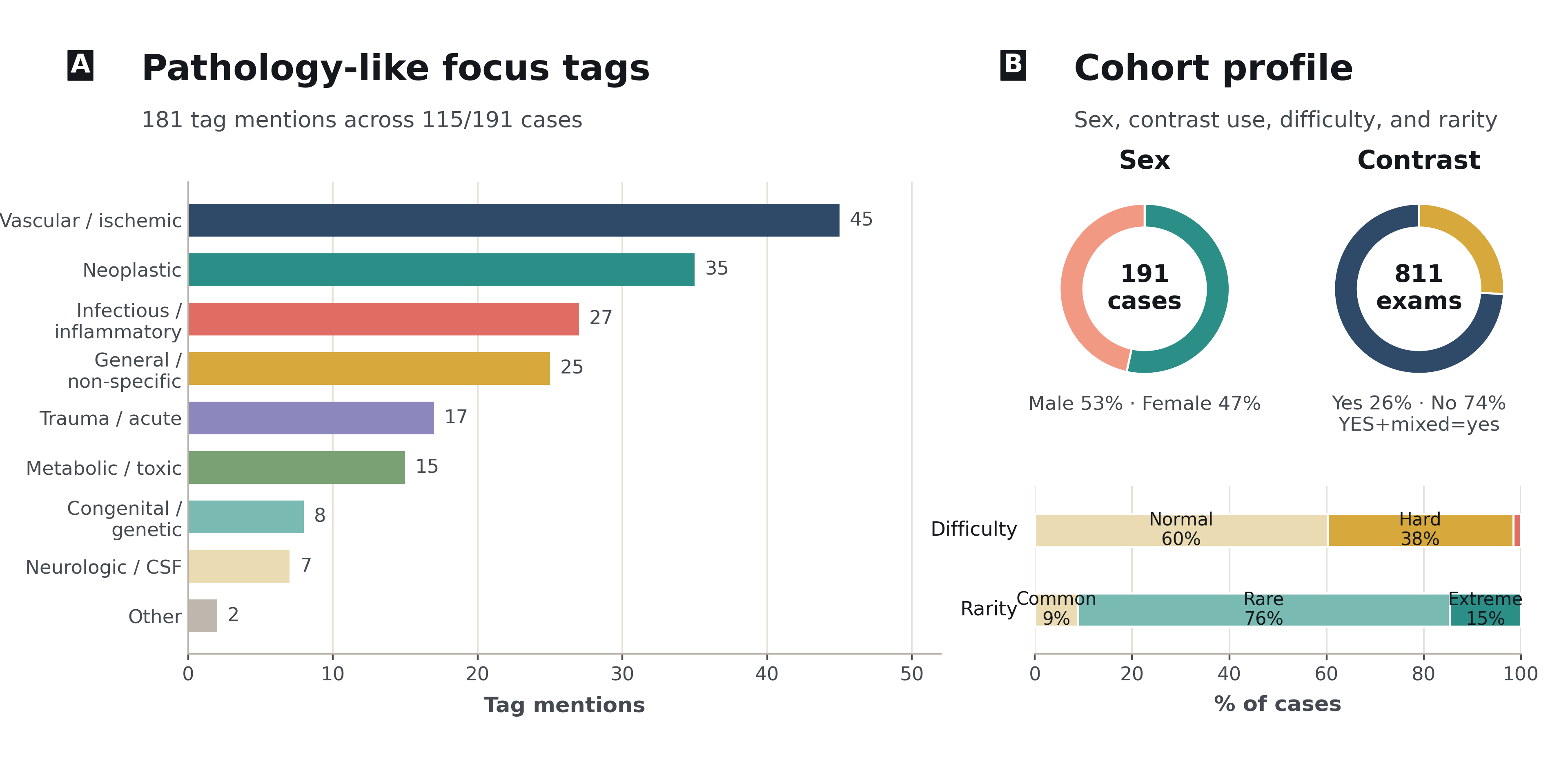}
    \caption{Distribution of the official release/evaluation cases across the main dataset attributes, including pathology/tag composition and case-level metadata such as rarity and difficulty. Pathology/tag counts are not necessarily mutually exclusive; rarity and difficulty percentages use the 191-case release/evaluation set as denominator. The exam/contrast metadata panel uses the 811 imaging-examination records before route-pool filtering.}
    \label{fig:data_distribution_part1}
\end{figure}

\begin{figure}[!htbp]
    \centering
    \includegraphics[width=1\linewidth]{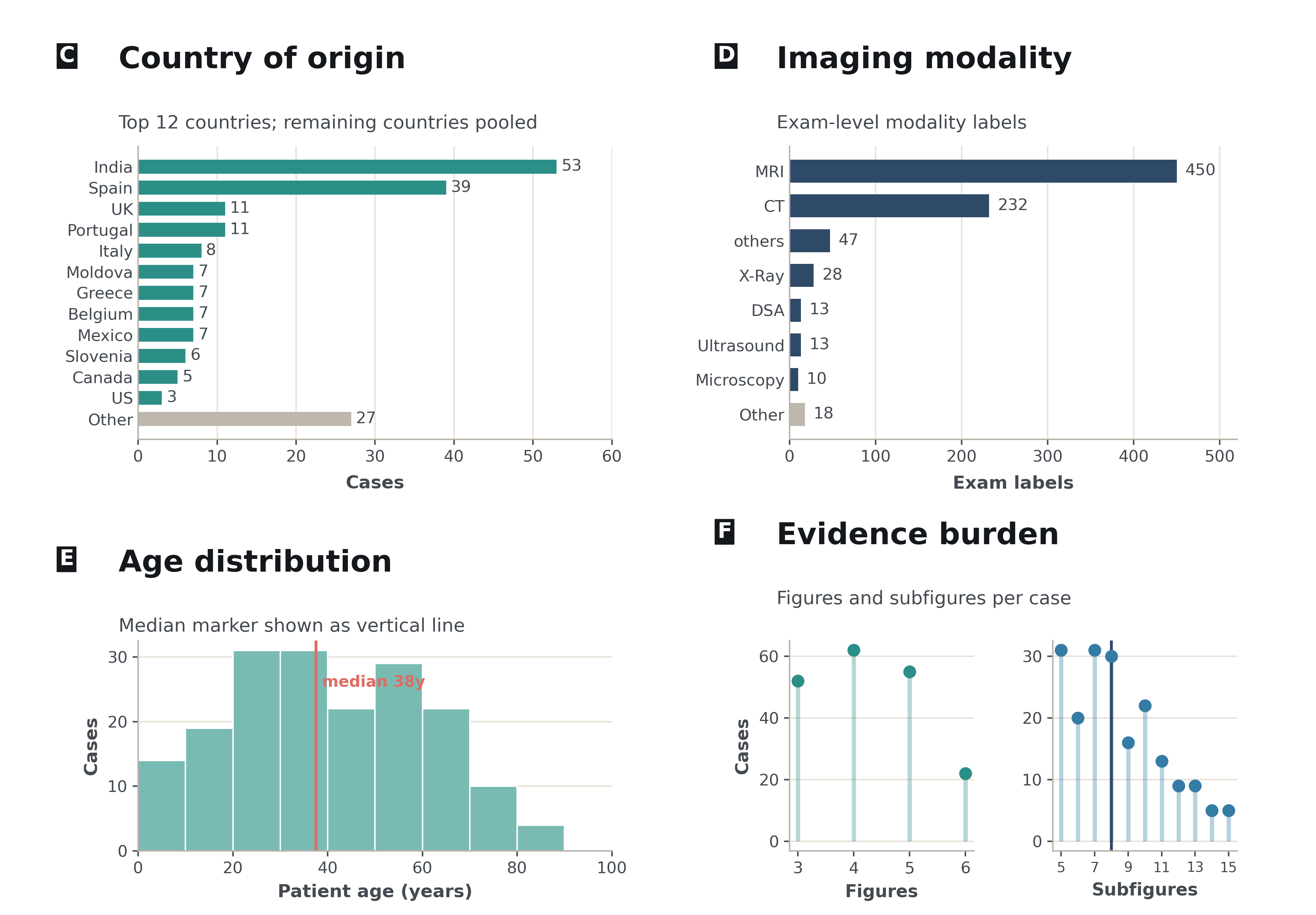}
    \caption{Distribution of imaging evidence and related benchmark characteristics, including case origin, patient age distribution, modality and sequence usage, exam counts, and image/subfigure counts across the 191-case release/evaluation set. The exam-count panels correspond to the 811 imaging-examination metadata records in the release export; official route/request metrics use the 785-record route-evaluable subset after excluding future follow-up records.}
    \label{fig:data_distribution_part2}
\end{figure}

This section summarizes the composition of \bench{} and provides a compact view of its case and evidence distributions. The annotation pipeline initially covered 211 candidate cases. After release filtering, the official release/evaluation set contains 191 cases, 811 imaging-examination metadata records, 789 diagnostic-step annotations, and 1,609 image/subfigure records, as recorded in the released nested JSON and Croissant metadata. Of the 811 imaging-examination records, 26 records are marked as future follow-up by numeric \texttt{time\_past < 0} and are excluded from the official route/request metrics, yielding 785 requestable route-evaluable evidence units. The core dataset statistics are reported in Table~\ref{tab:dataset_stats}. Fig.~\ref{fig:data_distribution_part1} visualizes the distribution of the released benchmark cases across the main dataset attributes, while Fig.~\ref{fig:data_distribution_part2} presents the complementary distribution of imaging evidence and related data characteristics.

\begin{table}[!htbp]
\centering
\small
\setlength{\tabcolsep}{5pt}
\renewcommand{\arraystretch}{1.15}
\begin{tabular*}{\linewidth}{@{\extracolsep{\fill}}>{\raggedright\arraybackslash}p{0.72\linewidth}>{\centering\arraybackslash}p{0.20\linewidth}@{}}
\toprule
\textbf{Statistic} & \textbf{Value} \\
\midrule
Annotated candidate cases & 211 \\
Excluded after release filtering & 20 \\
Official release/evaluation cases & 191 \\
Clinical domain & Neuroradiology \\
Cases with $\geq 2$ physician annotations & 211 \\
Retained cases with meta-review / final scoring & 191 \\
Imaging-examination metadata records in release export & 811 \\
Future-follow-up records excluded from route metrics & 26 \\
Official route-evaluable/requestable evidence units & 785 \\
Patient-provided prior/comparison records with numeric \(\texttt{time\_past > 0}\) & 6 \\
Diagnostic-step annotations & 789 \\
Image/subfigure records & 1,609 \\
Median exams per released case & 4 \\
Median images per released case & 8 \\
Distinct non-null modality strings & 28 \\
Rare released cases & 146 rare + 28 extreme rare (91.1\%) \\
Hard released cases & 73 hard + 3 extreme hard (39.8\%) \\
\bottomrule
\end{tabular*}
\caption{\textbf{Core benchmark statistics.} Summary of dataset size, annotation coverage, evidence volume, modality diversity, and case difficulty for \bench{}. Candidate-case counts refer to the annotation pipeline; image, rarity, and difficulty counts refer to the official 191-case release/evaluation set; route-evaluable/requestable evidence units exclude future follow-up records from the exam-metadata export.}
\label{tab:dataset_stats}
\end{table}

\clearpage
\section{Extended Result Analysis}

\label{app:extended_results}

\subsection{Ablation studies and oracle analysis}
\label{app:detailed_oracle_analysis}

This subsection reports representative per-model ablations to identify which components of the benchmark most strongly limit performance. In the ablation subset, average performance improves when the full image set is provided at once and improves further when oracle text findings are provided in place of raw image interpretation, although individual models can degrade under specific passive settings.

The ablations therefore point to bottlenecks that are not primarily the final answer format itself. Instead, current systems are constrained by the need to plan a multi-step workup, actively acquire the right evidence, and extract clinically useful information from raw images. Put differently, removing the acquisition problem or converting visual evidence into structured textual findings often raises model performance, indicating that planning, active evidence acquisition, and visual reasoning remain important failure modes.

\begin{table*}[!htbp]
    \centering
    \scriptsize
    \setlength{\tabcolsep}{3pt}
    \begin{tabular}{llcccccc}
        \toprule
        Model & Setting & \(S_{\text{dx}}\) & \(S_{\text{loc}}\) & \(S_{\text{ER}}\) & \(S_{\text{order}}\) & \(S_{\text{traj}}\) & Overall Change \\
        \midrule
        \multirow{6}{*}{Gemini 3 Flash} & Default setting & 0.54 & 0.58 & 0.76 & 0.61 & 0.49 & -- \\
        \cmidrule(lr){2-8}
        & History-only & 0.35\down{35.2} & 0.35\down{39.7} & -- & -- & 0.35\down{28.6} & \down{34.5} \\
        & All-images-at-once & 0.60\up{11.1} & 0.61\up{5.2} & -- & -- & 0.60\up{22.4} & \up{12.9} \\
        & Random-order reveal & 0.56\up{3.7} & 0.57\down{1.7} & -- & -- & 0.50\up{2.0} & \up{1.3} \\
        & Gold-order reveal & 0.58\up{7.4} & 0.59\up{1.7} & -- & -- & 0.53\up{8.2} & \up{5.8} \\
        & Oracle findings & 0.66\up{22.2} & 0.72\up{24.1} & 0.83\up{9.2} & 0.62\up{1.6} & 0.59\up{20.4} & \up{15.5} \\
        \midrule
        \multirow{6}{*}{Gemini 3.1 Pro} & Default setting & 0.54 & 0.57 & 0.72 & 0.57 & 0.50 & -- \\
        \cmidrule(lr){2-8}
        & History-only & 0.35\down{35.2} & 0.35\down{38.6} & -- & -- & 0.34\down{32.0} & \down{35.3} \\
        & All-images-at-once & 0.60\up{11.1} & 0.59\up{3.5} & -- & -- & 0.60\up{20.0} & \up{11.5} \\
        & Random-order reveal & 0.60\up{11.1} & 0.62\up{8.8} & -- & -- & 0.53\up{6.0} & \up{8.6} \\
        & Gold-order reveal & 0.60\up{11.1} & 0.60\up{5.3} & -- & -- & 0.55\up{10.0} & \up{8.8} \\
        & Oracle findings & 0.71\up{31.5} & 0.70\up{22.8} & 0.79\up{9.7} & 0.57\up{0.0} & 0.62\up{24.0} & \up{17.6} \\
        \midrule
        \multirow{6}{*}{Claude Sonnet 4.6} & Default setting & 0.42 & 0.51 & 0.71 & 0.54 & 0.39 & -- \\
        \cmidrule(lr){2-8}
        & History-only & 0.30\down{28.6} & 0.36\down{29.4} & -- & -- & 0.30\down{23.1} & \down{27.0} \\
        & All-images-at-once & 0.46\up{9.5} & 0.54\up{5.9} & -- & -- & 0.46\up{17.9} & \up{11.1} \\
        & Random-order reveal & 0.46\up{9.5} & 0.55\up{7.8} & -- & -- & 0.40\up{2.6} & \up{6.6} \\
        & Gold-order reveal & 0.47\up{11.9} & 0.55\up{7.8} & -- & -- & 0.43\up{10.3} & \up{10.0} \\
        & Oracle findings & 0.69\up{64.3} & 0.82\up{60.8} & 0.80\up{12.7} & 0.53\down{1.9} & 0.59\up{51.3} & \up{37.4} \\
        \midrule
        \multirow{6}{*}{Qwen3.5-35B-A3B} & Default setting & 0.28 & 0.47 & 0.42 & 0.19 & 0.27 & -- \\
        \cmidrule(lr){2-8}
        & History-only & 0.18\down{35.7} & 0.27\down{42.6} & -- & -- & 0.18\down{33.3} & \down{37.2} \\
        & All-images-at-once & 0.32\up{14.3} & 0.51\up{8.5} & -- & -- & 0.32\up{18.5} & \up{13.8} \\
        & Random-order reveal & 0.37\up{32.1} & 0.51\up{8.5} & -- & -- & 0.33\up{22.2} & \up{21.0} \\
        & Gold-order reveal & 0.39\up{39.3} & 0.56\up{19.1} & -- & -- & 0.35\up{29.6} & \up{29.3} \\
        & Oracle findings & 0.49\up{75.0} & 0.66\up{40.4} & 0.56\up{33.3} & 0.32\up{68.4} & 0.43\up{59.3} & \up{55.3} \\
        \midrule
        \multirow{6}{*}{MedGemma 27B} & Default setting & 0.16 & 0.26 & 0.57 & 0.44 & 0.16 & -- \\
        \cmidrule(lr){2-8}
        & History-only & 0.14\down{12.5} & 0.23\down{11.5} & -- & -- & 0.14\down{12.5} & \down{12.2} \\
        & All-images-at-once & 0.14\down{12.5} & 0.27\up{3.8} & -- & -- & 0.14\down{12.5} & \down{7.1} \\
        & Random-order reveal & 0.18\up{12.5} & 0.30\up{15.4} & -- & -- & 0.18\up{12.5} & \up{13.5} \\
        & Gold-order reveal & 0.18\up{12.5} & 0.29\up{11.5} & -- & -- & 0.18\up{12.5} & \up{12.2} \\
        & Oracle findings & 0.33\up{106.2} & 0.54\up{107.7} & 0.69\up{21.1} & 0.52\up{18.2} & 0.29\up{81.2} & \up{66.9} \\
        \midrule
        \multirow{6}{*}{Average} & Default setting & 0.39 & 0.48 & 0.64 & 0.47 & 0.36 & -- \\
        \cmidrule(lr){2-8}
        & History-only & 0.26\down{32.0} & 0.31\down{34.7} & -- & -- & 0.26\down{27.6} & \down{31.4} \\
        & All-images-at-once & 0.42\up{9.3} & 0.50\up{5.4} & -- & -- & 0.42\up{17.1} & \up{10.6} \\
        & Random-order reveal & 0.43\up{11.9} & 0.51\up{6.7} & -- & -- & 0.39\up{7.2} & \up{8.6} \\
        & Gold-order reveal & 0.44\up{14.4} & 0.52\up{8.4} & -- & -- & 0.41\up{12.7} & \up{11.8} \\
        & Oracle findings & 0.58\up{48.5} & 0.69\up{43.9} & 0.73\up{15.4} & 0.51\up{8.9} & 0.50\up{39.2} & \up{31.2} \\
        
        \bottomrule
    \end{tabular}
    \caption{\textbf{Per-model bottleneck ablations and oracle analyses.} In this representative ablation subset, average performance improves when all images are revealed at once and improves further when oracle text findings are provided, indicating that passive access to evidence and textualized findings can substantially reduce task difficulty. Individual models may degrade under specific passive settings. Passive settings expose fixed evidence rather than model-requested exams, so route and request metrics are structurally undefined and shown as ``--''. Overall Change reports the mean relative change over available metrics with respect to each model's default setting.}
    \label{tab:ablations}
\end{table*}

\begin{figure}[!htbp]
    \centering
    \includegraphics[width=1\linewidth]{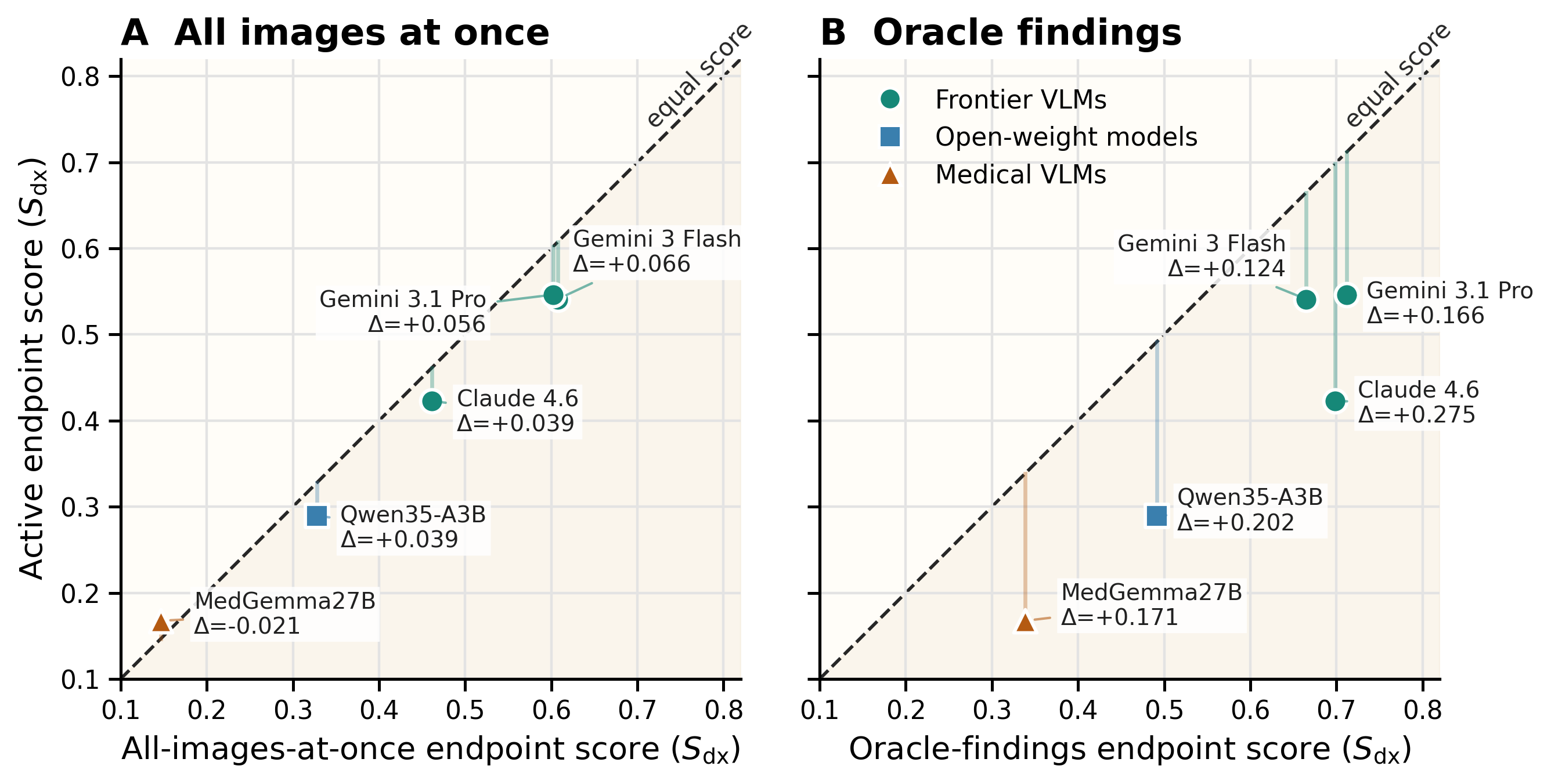}
    \caption{\textbf{Benchmark gap under passive evaluation with $S_{\mathrm{dx}}$ as the metric.} When models are given all images at once or oracle text findings instead of having to actively request and interpret evidence, their apparent performance improves. These passive settings remove the need for clinically grounded evidence acquisition, planning, and intermediate belief updating, and therefore obscure the gap between strong surface-level accuracy and practical diagnostic trajectory quality.}
    \label{fig:benchmark_gap}
\end{figure}

\begin{figure}[!htbp]
    \centering
    \includegraphics[width=1\linewidth]{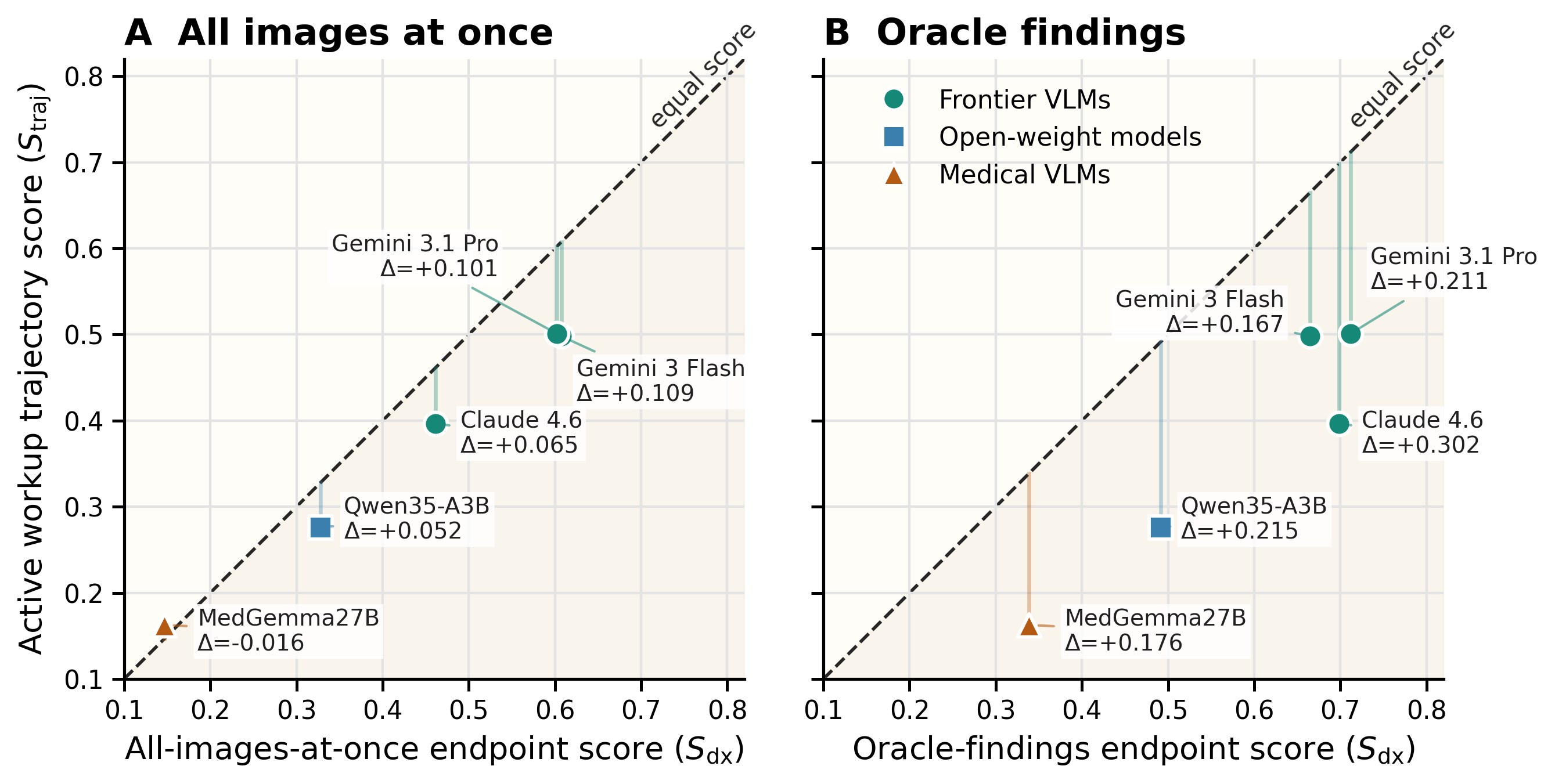}
    \caption{\textbf{Benchmark gap comparing passive endpoint score and active trajectory score.} This panel compares passive endpoint diagnosis score ($S_{\mathrm{dx}}$) against active-workup trajectory score ($S_{\mathrm{traj}}$). It should therefore be interpreted as a cross-metric diagnostic-process comparison, not as a within-metric passive-versus-active ablation. Passive settings remove the need for clinically grounded evidence acquisition, planning, and intermediate belief updating, and can obscure the gap between endpoint accuracy and practical diagnostic trajectory quality.}
    \label{fig:benchmark_gap_2}
\end{figure}

\clearpage
\subsection{Efficiency--accuracy tradeoff}
\label{sec:eff-acc}

\begin{figure}[th]
    \centering
    \includegraphics[width=\linewidth]{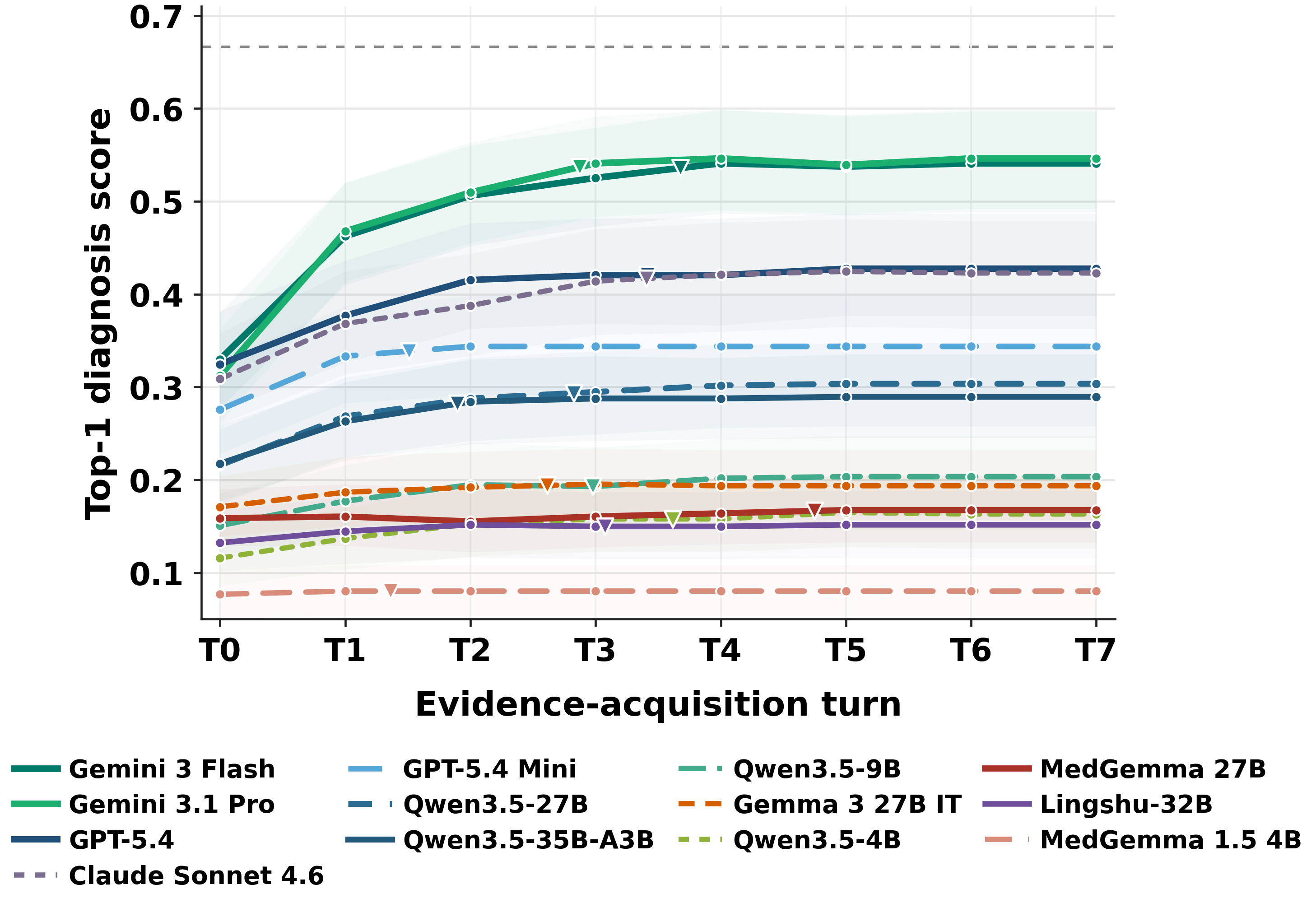}
    \caption{\textbf{Diagnosis trajectory over turns.} We track the normalized top-1 diagnosis score as models acquire sequential evidence. Frontier models improve more consistently with additional evidence, whereas open-weight and medical/radiology-adapted models show smaller or less stable gains. Shaded bands indicate 95\% bootstrap confidence intervals; triangles mark mean stopping turns.}
    \label{fig:trajectory}
\end{figure}

Fig.~\ref{fig:trajectory} shows that sequential evidence can improve endpoint performance, especially for frontier models, but endpoint improvement and clinical sufficiency are not equivalent. Some models stop early or request little without completing the workup; others continue requesting evidence without translating it into better diagnoses. The desired behavior is therefore not simply fewer or more requests, but a correct localized diagnosis supported by essential evidence and obtained without invalid or low-value examinations.

\begin{figure}[!htbp]
    \centering
    \includegraphics[width=0.8\linewidth]{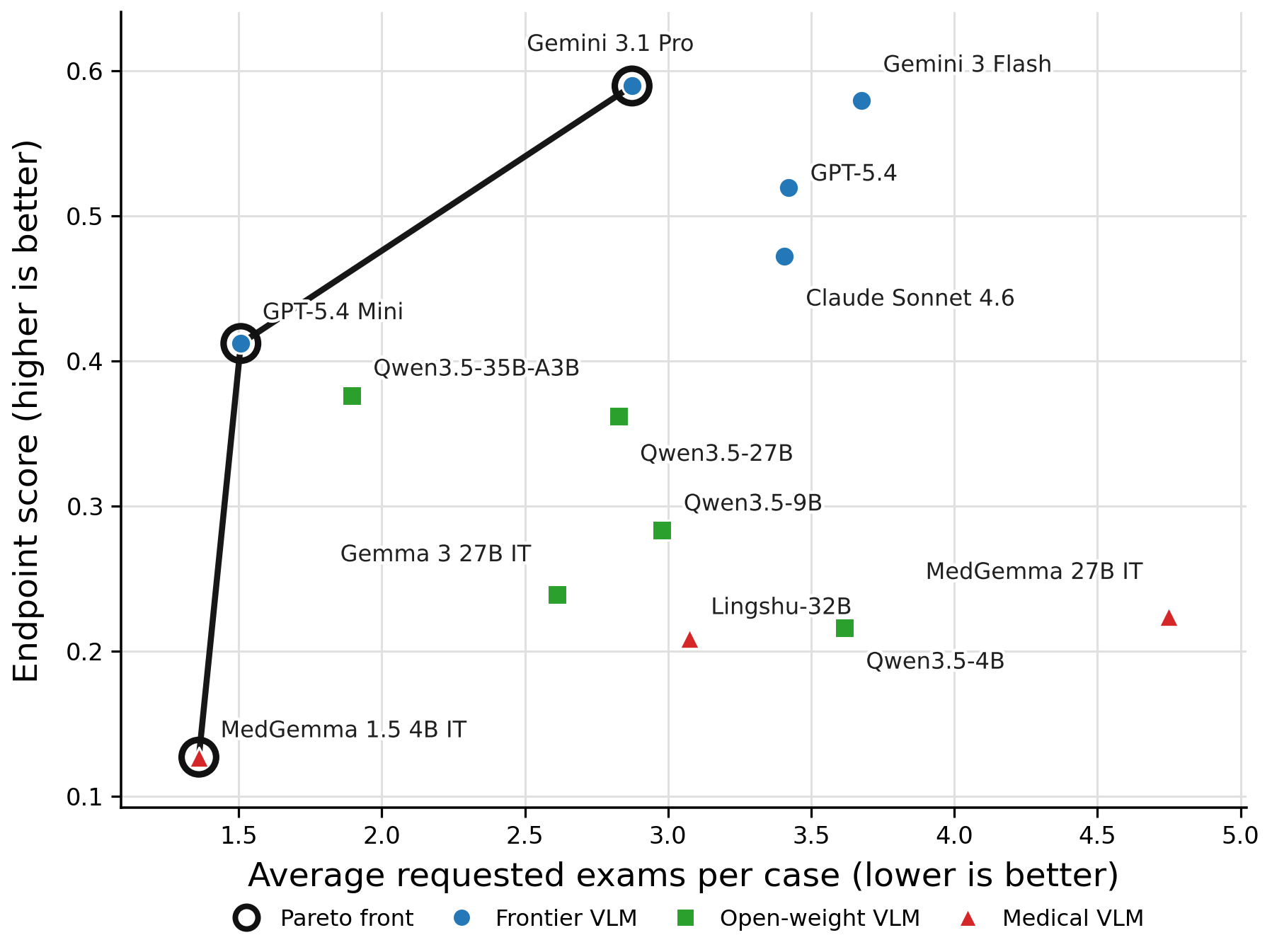}
    \caption{\textbf{Efficiency--accuracy Pareto frontier.} Endpoint diagnostic performance ($S_{\mathrm{dx}}$) versus the number of requested exams. MedGemma 1.5 4B, GPT-5.4 Mini, and Gemini 3.1 Pro lie on the plotted Pareto frontier: Gemini 3.1 Pro achieves the strongest diagnostic performance among these frontier points but requires relatively more exams, whereas GPT-5.4 Mini occupies a lower-request point on the same frontier.}
    \label{fig:pareto}
\end{figure}

Figure~\ref{fig:pareto} summarizes the relationship between endpoint quality and request efficiency. Frontier models occupy the high-accuracy region but also require more requests. Very short trajectories are not necessarily preferable, since some models stop early with relatively low final diagnosis scores. Conversely, longer trajectories do not guarantee better workup quality, as some models (e.g., MedGemma 27B) request the most examinations without corresponding gains.

This tradeoff motivates decomposed reporting. A model that requests few exams may be efficient, premature, or unable to formulate useful requests. A model that requests many exams may be thorough, over-testing, or stuck in uncertainty. Table~\ref{tab:main_results} distinguishes these cases by jointly reporting essential recall, optional burden, unmatched-request rate, endpoint quality, and timing metrics.

\subsection{Slice analysis}
To test whether model failures are primarily driven by intrinsically harder cases, we stratify performance by physician-annotated rarity and difficulty. These labels reflect how unusual or challenging the cases are for human experts, and therefore provide a natural check on whether current models fail most severely on the same subsets that are difficult for clinicians.

\begin{figure}[!htbp]
    \centering
    \includegraphics[width=1\linewidth]{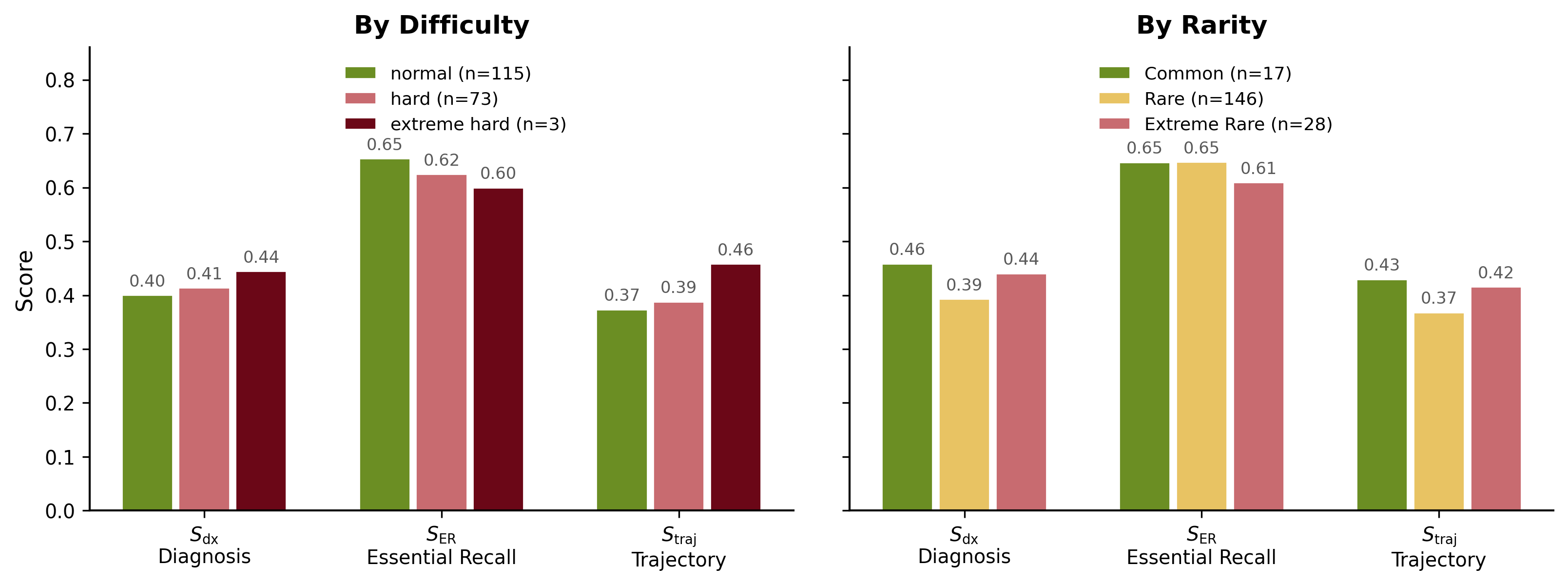}
    \caption{\textbf{Slice analysis by physician-rated rarity and difficulty.} Bars report the mean performance across five representative models: Gemini 3.1 Pro, Gemini 3 Flash, GPT-5.4, GPT-5.4 Mini, and MedGemma 27B. Model performance remains relatively stable across cases annotated as more common versus rarer and easier versus harder for human experts. The extreme-hard and common-rarity slices are small and should be interpreted descriptively. The weak dependence on these slices suggests that performance bottlenecks are driven less by case-level difficulty itself than by limited planning and active evidence-acquisition ability.}
    \label{fig:slice_analysis}
\end{figure}

Figure~\ref{fig:slice_analysis} shows that this is not the dominant pattern in our benchmark: model performance does not decrease markedly on cases labeled as rarer or more difficult by physicians. This weak slice dependence suggests that the main bottleneck is not simply visual or diagnostic difficulty at the case level. Instead, the broader results above are more consistent with a different explanation: current models struggle because they lack strong planning and active evidence-acquisition capabilities, which are required across the benchmark, including in cases that are not exceptionally rare or difficult for humans. In this sense, \bench{} exposes a process failure rather than only a case-complexity failure.

\subsection{Confidence alignment and calibration}
\label{sec:confidence}

\paragraph{Additional metrics.} Because the four probabilities of the differential diagnosis are normalized to sum to 1, mean raw confidence is not an informative summary by itself. For calibration plots, we therefore use the final top-1 probability \(p_{\max}\) as the scalar confidence. A simple Brier-style calibration objective is
\begin{equation}
\mathrm{Brier}_{\mathrm{top1}}
=
\frac{1}{N}\sum_{n=1}^{N}(p_{\max,n} - S_{\mathrm{dx},n})^2,
\label{eq:brier}
\end{equation}
where the normalized ordinal diagnosis score serves as the target; this is a Brier-style objective rather than a standard binary Brier score. In the experiments, we report \(S_{\mathrm{conf}}\), \(\mathrm{Brier}_{\mathrm{top1}}\), and reliability plots based on \(p_{\max}\).

The confidence metrics reveal that several models assign substantial probability mass to unmatched or clinically unacceptable diagnoses. Gemini 3 Flash and Gemini 3.1 Pro have the strongest final confidence alignment scores (\(S_{\mathrm{conf}}=0.531\) and \(0.504\)), whereas several open-weight and medical/radiology-adapted models have negative confidence-alignment scores. This means that, even when such models include a plausible diagnosis in the final differential, they often distribute confidence poorly across alternatives.

\begin{figure}[!htbp]
    \centering
    \includegraphics[width=1\linewidth]{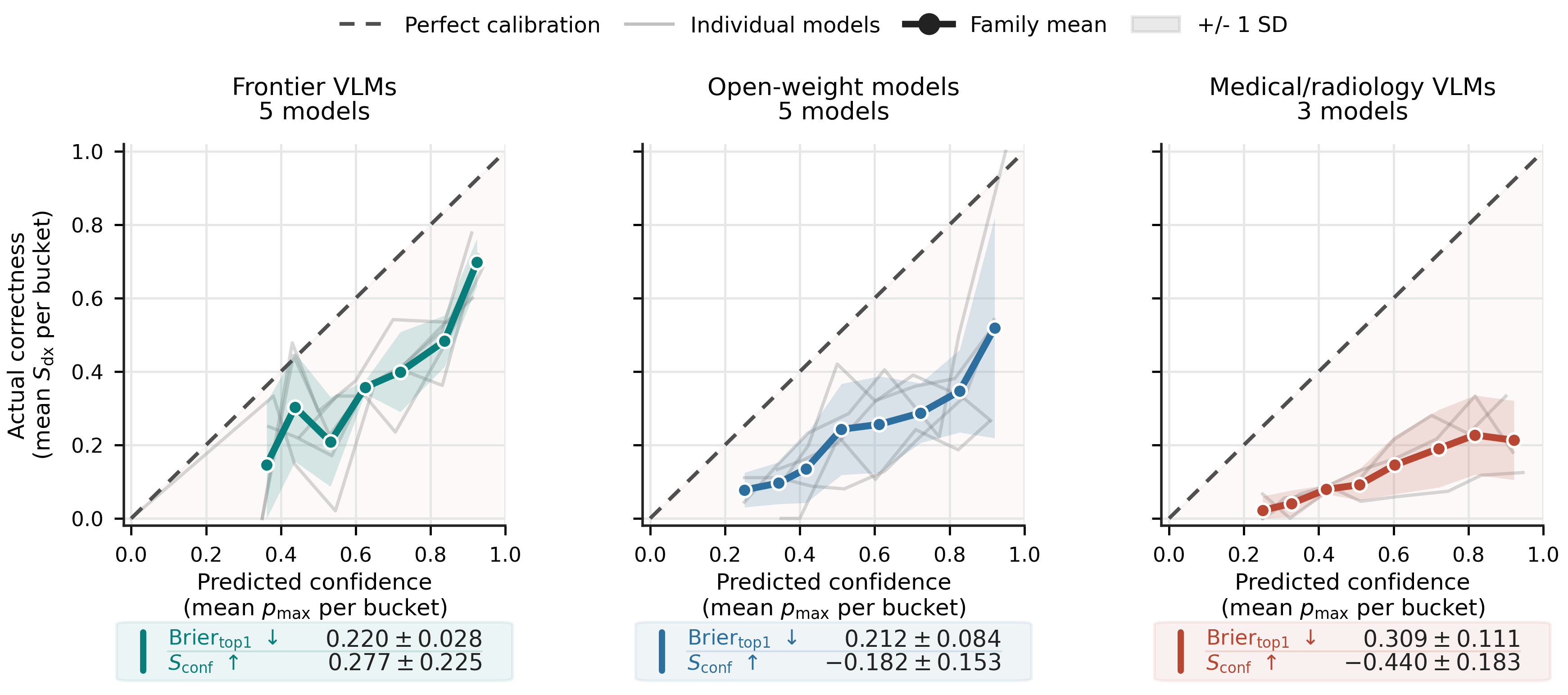}
    \caption{\textbf{Calibration analysis overview.} Reliability diagrams for frontier models, open-weight models, and medical/radiology-adapted models using the final top-1 probability against the normalized diagnosis score ($S_{\mathrm{dx}}$).}

    \label{fig:calibration_analysis}
\end{figure}

This analysis is useful because the benchmark evaluates a four-way differential rather than only a single label. A model that includes the correct answer but assigns high confidence to implausible alternatives is not equivalent to a model that concentrates probability mass on the correct diagnosis and clinically plausible differentials. Figure~\ref{fig:calibration_analysis} reports reliability plots based on the final top-1 probability in frontier models, open-weight models, and medical/radiology-adapted models.

\newpage
\section{Annotation Details and Quality}
\label{app:annotation_quality}

To initialize the benchmark schema, we first used an LLM to draft a structured template for each EuroRad case. This draft proposed a normalized organization for patient history, exam bundles, figure-level evidence units, differential diagnoses, and rubric placeholders. The draft was then reviewed and corrected by physicians. In particular, physicians audited modality naming, acquisition type, anatomical view, imaged region, temporal context, contrast usage, diagnostic-step wording, and rubric formulations.

The released annotation scheme contains three step-importance labels: \texttt{essential}, \texttt{optional}, and \texttt{unnecessary}. Preferred order is defined over requestable exam bundles rather than over raw images. Difficulty and rarity are case-level labels. In the annotation pipeline, 211 candidate cases received at least two physician annotations, yielding 430 physician annotations in total. After release filtering, 191 cases are retained in the official release/evaluation set. Agreement statistics from the annotation logs are: 75.2\% exact agreement on step importance (\(\kappa = 0.360\)), 74.1\% pairwise agreement on preferred order, 61.3\% exact agreement on difficulty (\(\kappa = 0.267\)), and 67.3\% exact agreement on rarity (\(\kappa = 0.187\)). We also observe that 137 out of 211 cases required correction of LLM template artifacts.

Disagreement patterns were clinically interpretable. For step importance, most conflicts were \texttt{essential} versus \texttt{optional}. For preferred order, most conflicts were adjacent-rank swaps rather than large order reversals. For difficulty and rarity, the most common disagreements were \texttt{normal} versus \texttt{hard} and \texttt{rare} versus \texttt{extreme rare}, respectively. These disagreements were resolved through the meta-review pipeline, yielding one final label set aligned to the official exam-bundle action space.

\begin{table}[!htbp]
\centering
\small
\setlength{\tabcolsep}{5pt}
\renewcommand{\arraystretch}{1.15}
\begin{tabular*}{\linewidth}{@{\extracolsep{\fill}}>{\raggedright\arraybackslash}p{0.74\linewidth}>{\centering\arraybackslash}p{0.18\linewidth}@{}}
\toprule
\textbf{Annotation quantity / quality metric} & \textbf{Value} \\
\midrule
Total physician annotations & 430 \\
Mean annotations per candidate case & 2.04 \\
Agreement on step importance & 75.2\% (\(\kappa=0.360\)) \\
Agreement on preferred order & 74.1\% pairwise \\
Agreement on rarity & 67.3\% (\(\kappa=0.187\)) \\
Agreement on difficulty & 61.3\% (\(\kappa=0.267\)) \\
Cases requiring adjudication & 182/211 \\
LLM template artifacts corrected by physicians & 137/211 \\
\bottomrule
\end{tabular*}
\caption{\textbf{Annotation quality and adjudication summary.} Overview of annotation volume, inter-annotator agreement, adjudication frequency, and physician correction of template artifacts across the 211-case candidate annotation pool and the retained release/evaluation set.}
\label{tab:annotation_quality}
\end{table}

\begin{table*}[!htbp]
    \centering
    \scriptsize
    \setlength{\tabcolsep}{4pt}
    \begin{tabular}{p{0.12\linewidth}p{0.10\linewidth}p{0.13\linewidth}p{0.13\linewidth}p{0.17\linewidth}p{0.11\linewidth}p{0.09\linewidth}}
        \hline
        Label family & Annotator 1 vs 2 & Agreement metric & After adjudication & Common disagreement type & Resolved by meta-review? & Notes \\
        \hline
        Step importance & 75.2\% & Exact agreement; \(\kappa=0.360\) & Final label in JSON & Essential vs.\ optional & Yes & \(n=814\) step pairs \\
        Exam order & 74.1\% & Pairwise order concordance & Final preferred order & Adjacent-rank swaps & Yes & \(n=1492\) exam pairs \\
        Difficulty & 61.3\% & Exact agreement; \(\kappa=0.267\) & Final difficulty label & Normal vs.\ hard & Yes & 76 conflict cases \\
        Rarity & 67.3\% & Exact agreement; \(\kappa=0.187\) & Final rarity label & Rare vs.\ extreme rare & Yes & 63 conflict cases \\
        \hline
    \end{tabular}
    \caption{\textbf{Agreement and adjudication breakdown.} Detailed agreement statistics by label family, including the metric used, the predominant source of disagreement, and whether conflicts were resolved during meta-review. Pair counts in the notes column are annotation-pair counts and are not the same quantity as the 789 final diagnostic-step annotation records in the release export.}
    \label{tab:appendix_agreement}
\end{table*}

\clearpage

\section{Additional Task-Protocol Details}
\label{app:protocol_details}

\paragraph{Request matching.}
The official scored evidence unit is an imaging exam bundle. Each turn permits at most one requested examination. The request resolver compares the free-form request against the hidden requestable exam inventory and returns at most one matched bundle. A matched request reveals the corresponding image bundle and minimal exam metadata. An unmatched request reveals no evidence and is retained in the trajectory for unmatched-request scoring. This design preserves open-ended natural-language requests while keeping route scoring auditable at the bundle level.

\paragraph{Output validation.}
The official turn-level schema contains four fields: \texttt{action}, \texttt{requested\_examination} when applicable, \texttt{current\_differential}, and \texttt{final\_location} on the stop turn. The differential diagnosis must contain exactly four diagnosis-probability pairs, and the probabilities must sum to 1. The final localization is represented structurally rather than as free text, using components such as laterality, anatomical region, and, when appropriate, more specific substructures.

\paragraph{Gold route labels.}
The annotation schema stores figure-linked diagnostic steps and exam-level order metadata. Each diagnostic step is mapped to a requestable imaging exam bundle through its associated figure identifier. In the current release, each annotated figure appears at most once in \texttt{diagnostic\_steps} and therefore carries at most one final importance label. Exam bundles without explicit step annotations are treated as optional in the official route metrics. Because several cases contain tied preferred-order stages, route order is evaluated as a partial order with ties rather than a strict total ranking. Records marked as future follow-up by numeric \texttt{time\_past < 0} are excluded from the official request pool and from route metrics, even when they remain in the full case record.

\paragraph{Efficiency reporting.}
The request budget \(B=6\) is a benchmark hyperparameter, not a claim about clinical practice. It is intended to allow multi-step evidence acquisition while exposing premature stopping, repeated unmatched requests, and inefficient workups. For \(T_{\mathrm{guess}}\) and \(T_{\mathrm{clin}}\), cases that do not satisfy the required condition within the budget are recorded as unreached. We therefore report both timing summaries and the fraction of reached cases when analyzing diagnostic efficiency.

\clearpage

\section{Scorer Implementation and Judge Reliability}
\label{app:judge_reliability}

Route metrics are fully deterministic after request matching. Endpoint scoring and trajectory-category labeling are produced by a rubric-conditioned LLM-as-a-Judge call with structured JSON output. The primary automatic judge in the reported full runs is \textbf{Gemini 3 Flash}; judge decoding uses temperature 0. Candidate-model decoding settings are reported separately in Appendix~\ref{app:official_agent_prompt}. For each case, the judge receives the case-specific diagnosis and localization rubrics, the model's final output, and the set of unique diagnosis strings that appeared anywhere in the trajectory. It returns normalized endpoint scores for diagnosis and localization, together with exact-match, acceptable-differential, or unmatched labels for trajectory diagnoses.

The scorer is run with fixed prompts and versioned parsing code so that identical model traces produce identical parsed score outputs, conditional on the judge response. We use Gemini 3 Flash as the primary judge because it provides scalable rubric-based evaluation for open-ended diagnostic answers and shows moderate-to-high macro binary agreement with the human reference in Table~\ref{tab:judge_pairwise_agreement}. The rule-based scorer is retained as a deterministic reference, but it is less flexible for partially correct answers, especially in the neuroradiology cases. These cases usually have synonym-heavy diagnostic expressions and nuanced localization descriptions.

To assess whether benchmark conclusions are sensitive to judge choice, we conduct a judge ablation on a frozen set of model outputs from representative testee models sampled from the benchmark traces. The evaluated outputs include final diagnosis, ranked differential diagnosis, localization, and supporting evidence summaries. Judges are blinded to the identity of the testee model and assign scores using the same ordinal \(0\)--\(3\) rubrics used in the main benchmark. The judge set includes Gemini 3 Flash, GPT-5.4-mini, GPT-5.4, a rule-based scorer, and a human reference scorer.

Table~\ref{tab:judge_pairwise_agreement} reports pairwise macro binary agreement between judge pairs. The ordinal 0--3 scores are binarized into clinically unacceptable (0--1) versus clinically acceptable (2--3) before agreement is computed. This table therefore does not report quadratic weighted Cohen's kappa, exact ordinal agreement, or within-one ordinal agreement.

This analysis measures whether the primary LLM judge is closer to the available human reference than the rule-based scorer and checks whether benchmark rankings are robust to judge choice. In the main benchmark, we therefore report Gemini 3 Flash scores as the primary automatic scores and include judge-ablation results in this appendix.

\begin{table*}[!htbp]
\centering
\small
\begin{tabular}{lcccc}
\hline
Judge & GPT-5.4-mini & GPT-5.4 & Rule-based & Human reference \\
\hline
Gemini-Flash 
& 0.914 
& 0.940 
& 0.659 
& 0.741 \\

GPT-5.4-mini 
& -- 
& 0.921 
& 0.670 
& 0.702 \\

GPT-5.4 
& -- 
& -- 
& 0.669 
& 0.737 \\

Rule-based 
& -- 
& -- 
& -- 
& 0.648 \\
\hline
\end{tabular}
\caption{\textbf{Pairwise judge agreement.} Each cell reports macro binary agreement between two judges; this table does not report ordinal kappa or exact ordinal agreement. For each ordinal 0--3 rubric score, we first binarize the score into clinically unacceptable (0--1) versus clinically acceptable (2--3), so minor ordinal differences within the same acceptability band, such as 2 versus 3, are counted as agreement, whereas clinically meaningful disagreements, such as 0 versus 3, are counted as disagreement. Agreement is computed separately for diagnosis, localization, differential-list quality, and trajectory diagnosis scores, and the four agreement rates are then averaged with equal weight. The lower triangle is omitted because pairwise agreement is symmetric. The detailed LLM-as-judge prompt and rule-based judge criteria are provided in Section~\ref{app:prompt_matching_eval}.}
\label{tab:judge_pairwise_agreement}
\end{table*}

\newpage
\section{Annotation Interface}
\begin{figure}[!htbp]
    \centering
    \includegraphics[width=1\linewidth]{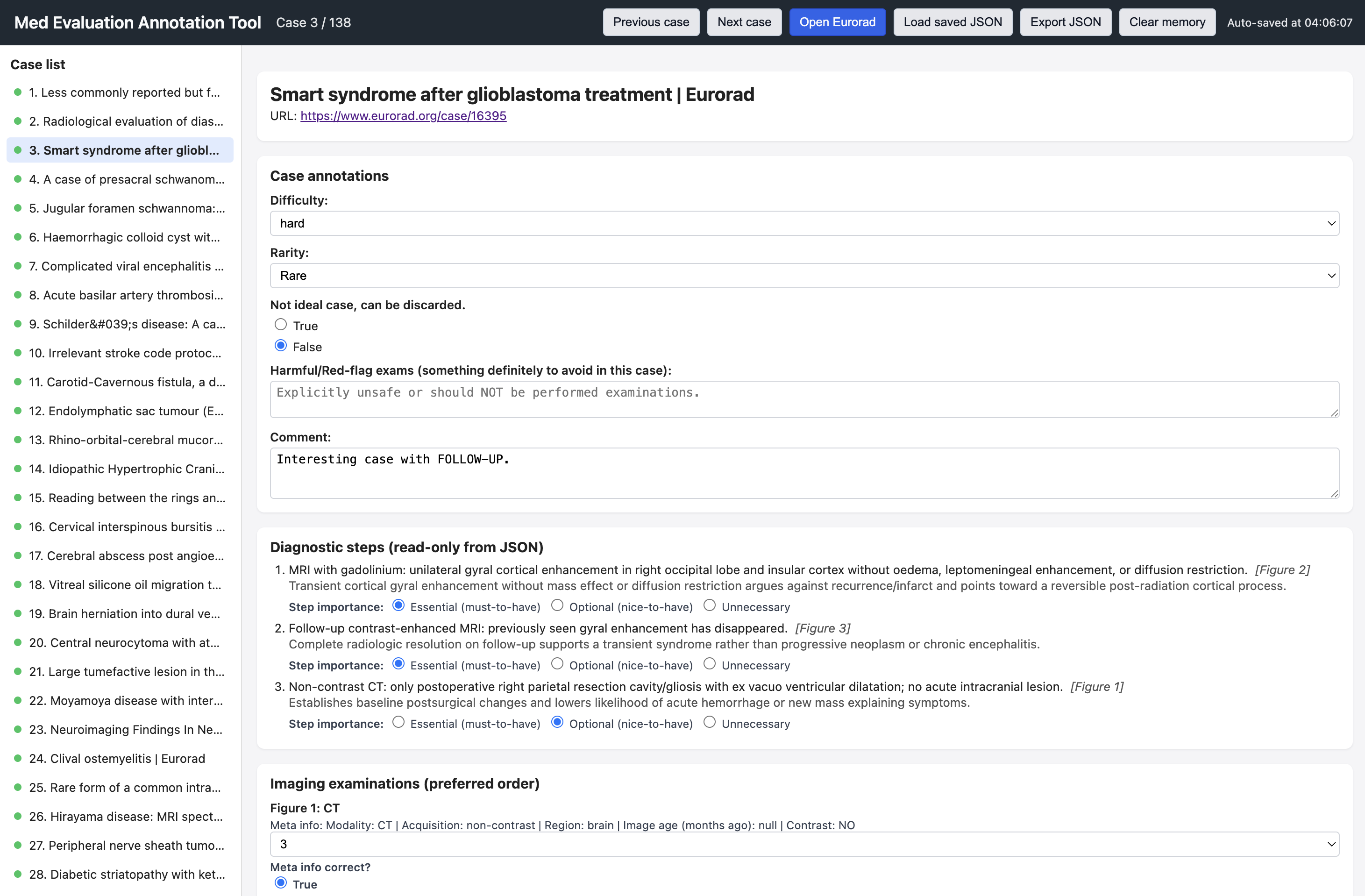}
    \caption{\textbf{Physician annotation interface.} Screenshot of the web-based review platform used to annotate case-level and exam-level benchmark metadata. Annotators label the importance of intermediate imaging steps, specify preferred exam order, assess case rarity and difficulty, and correct metadata or template artifacts, including modality, acquisition, view, imaged region, temporal context, contrast usage, and rubric inconsistencies.}
    \label{fig:annotation_interface}
\end{figure}

A screenshot of the annotation interface is illustrated in Fig. \ref{fig:annotation_interface}. The importance of intermediate imaging steps, a preferred order over imaging exams, case rarity and case difficulty are annotated via this platform. Further, physicians can correct exam-level metadata, including modality, acquisition, view, imaged region, temporal context, and contrast usage. Annotators also corrected template artifacts such as modality naming errors, weak rubric criteria, and mismatches between figures and structured fields.

\newpage
%

\lstdefinestyle{medprompt}{
  basicstyle=\ttfamily\scriptsize,
  breaklines=true,
  columns=fullflexible,
  keepspaces=true,
  showstringspaces=false,
  frame=single,
  framerule=0.25pt,
  linewidth=\linewidth,
  xleftmargin=0pt,
  xrightmargin=0pt,
  framexleftmargin=0.4em,
  framexrightmargin=0.4em,
  framextopmargin=0.25em,
  framexbottommargin=0.25em,
  aboveskip=0.6\baselineskip,
  belowskip=0.6\baselineskip
}

\providecommand{\promptblockheading}[1]{%
  \paragraph{#1}\leavevmode\par\nobreak\smallskip\noindent
}

\providecommand{\rubriclevel}[1]{\textbf{#1}}
\providecommand{\rubricitem}[2]{\item[\rubriclevel{#1}] #2}

\section{Evaluation Rubrics}
\label{app:evaluation_rubrics}

For each EuroRad case, we generated case-specific grading rubrics for the
open-ended diagnostic characterization task.  The rubric has two components:
\emph{Diagnosis}, which scores the most likely diagnosis on an ordinal 0--3
scale, and \emph{Localization}, which scores the anatomical precision of the
reported abnormality on the same 0--3 scale.  The exported JSON schema retains
the key \texttt{Localisation} for compatibility with the release manifest.  Each rubric also stores a
structured reference answer.  The diagnosis reference is a single string equal
to the final diagnosis.  The localisation reference is a three-level structure:
laterality, organ/region, and specific substructure or segment.

\paragraph{Rubric generation.}
Rubrics were generated offline from the scraped EuroRad case payload using the
rubric-generation script in \texttt{auto\_gpt\_part.py}.  The script calls the
OpenAI Responses API with JSON-schema structured output.  In the preprocessing
configuration used for this dataset, the rubric-generation model is
\texttt{gpt-5.2} unless explicitly overridden, with reasoning effort set to
\texttt{low}.  The model is instructed to use only the provided case payload and
not to browse.  This automated step was used only to create an initial rubric
draft: all generated rubrics were subsequently reviewed, corrected, and
finalized by human clinical experts before being used for evaluation.  The
diagnosis reference answer is then programmatically overwritten with
\texttt{raw\_case.final\_diagnosis} when available, ensuring that the rubric
reference diagnosis exactly matches the source case label.

\paragraph{Information provided to the rubric generator.}
The model receives one normalized \texttt{raw\_case} object.  This object
contains the scraped case metadata and clinical content, including the case
title, section, patient age/sex when available, clinical history, imaging
findings, discussion, final diagnosis, differential diagnosis, figure
information, image captions, and extracted image/figure details.  The actual
rubric-generation call does not use external information beyond this payload.

\promptblockheading{Rubric-generation prompt.}
\begin{lstlisting}[style=medprompt]
You are a senior radiology educator.

TASK
You will receive ONE EuroRad case payload under `raw_case` (already scraped). Using ONLY this content (no browsing),
generate a single rubric object for grading a 2-step open-ended response to the stem:
"Based on the imaging figures provided, what is the most likely diagnosis?"

The trainee's answer is expected to be structured in two parts:
1) Localisation -- where the abnormality is.
2) Diagnosis    -- the most likely diagnosis.

REQUIREMENTS
- Output must be a single JSON object that strictly matches the provided JSON Schema.
- Provide analytic criteria for score levels "3", "2", "1", "0" in EACH of the sections (Localisation / Diagnosis).
- Include a **reference_answer** for EACH section:
  - Localisation.reference_answer must be an object with three hierarchical steps:
    {"Laterality": "...", "Organ/Region": "...", "Specific Substructure/Segment": "..."}.
    The localisation criterion must explicitly evaluate precision along this hierarchy (laterality -> organ/region -> specific substructure/segment) without disclosing additional information beyond the rubric.
  - Diagnosis.reference_answer must equal the case's final diagnosis found in `raw_case.final_diagnosis` (verbatim; if unavailable, use "N/A").
- All rubric criteria must be concretely tailored to the specific EuroRad case content in `raw_case`. Avoid vague or generic wording; tie each scoring level to the actual anatomical structures, imaging modalities, and discriminative findings present in the case.
- Use exactly the key name **reference_answer** (do not use Key_Imaging_Findings).
- All output must be in English.

STYLE
- Concise, objective, image-centric language.
- Use neutral phrasing (e.g., "identifies precise compartment and segment" rather than giving the answer).

OUTPUT JSON SCHEMA (the tool enforces this):
{
  "type": "object",
  "additionalProperties": false,
  "properties": {
    "Localisation": {
      "type": "object",
      "additionalProperties": false,
      "properties": {
        "reference_answer": {
          "type": "object",
          "additionalProperties": false,
          "properties": {
            "Laterality": {"type":"string"},
            "Organ/Region": {"type":"string"},
            "Specific Substructure/Segment": {"type":"string"}
          },
          "required": ["Laterality","Organ/Region","Specific Substructure/Segment"]
        },
        "3": {"type":"string"}, "2": {"type":"string"}, "1": {"type":"string"}, "0": {"type":"string"}
      },
      "required": ["reference_answer","3","2","1","0"]
    },
    "Diagnosis": {
      "type": "object",
      "additionalProperties": false,
      "properties": {
        "reference_answer": {"type":"string"},
        "3": {"type":"string"}, "2": {"type":"string"}, "1": {"type":"string"}, "0": {"type":"string"}
      },
      "required": ["reference_answer","3","2","1","0"]
    }
  },
  "required": ["Localisation","Diagnosis"]
}
\end{lstlisting}

\subsection{Example Case-Specific Rubrics}
\label{app:example_case_rubrics}

The examples below show three case-specific rubrics used by the benchmark.
Each block lists the reference answer and the four ordinal scoring levels for both rubric components.

\subsubsection{Case 16266: Mixed pachygyria and subcortical band heterotopia}

\paragraph{Diagnosis reference.}
Mixed pachygyria and subcortical band heterotopia (grade 5 lissencephaly).

\begin{description}
\setlength{\itemsep}{0.25em}
\rubricitem{3}{Full diagnosis: mixed pachygyria with subcortical band heterotopia, within the lissencephaly--SBH spectrum, correctly identifying grade 5 lissencephaly.}
\rubricitem{2}{Identifies the key entity incompletely, such as subcortical band heterotopia or double cortex, but omits or misstates the grade.}
\rubricitem{1}{Related but non-specific or partial diagnosis, such as lissencephaly/pachygyria alone, neuronal migration disorder, or cortical malformation without specifying subcortical band heterotopia.}
\rubricitem{0}{Incorrect diagnosis inconsistent with the described double-cortex subcortical band and pachygyria, or no diagnosis provided.}
\end{description}

\paragraph{Localisation reference.}
Laterality: bilateral/symmetric. Organ/region: supratentorial cerebral
hemispheres. Specific substructure/segment: subcortical white matter
immediately deep to the cerebral cortex, forming a parallel ``double cortex''
band.

\begin{description}
\setlength{\itemsep}{0.25em}
\rubricitem{3}{Bilateral/symmetric supratentorial cerebral hemispheric process with a continuous subcortical band immediately deep to the cortex, producing a double-cortex appearance.}
\rubricitem{2}{Bilateral/symmetric cortical--subcortical abnormality in the cerebral hemispheres, but imprecise about the exact compartment.}
\rubricitem{1}{Broad brain/cortex/white-matter localisation without bilateral/symmetric distribution or subcortical band location.}
\rubricitem{0}{Incorrect or non-anatomical localisation, or no localisation.}
\end{description}

\subsubsection{Case 12789: Bilateral medial medullary infarction}

\paragraph{Diagnosis reference.}
Bilateral medial medullary infarction.

\begin{description}
\setlength{\itemsep}{0.25em}
\rubricitem{3}{Specific diagnosis of bilateral medial medullary infarction.}
\rubricitem{2}{Almost correct but missing a key qualifier, such as medial medullary infarction without bilateral involvement, or bilateral medullary infarction without medial distribution.}
\rubricitem{1}{Nonspecific or alternative brainstem diagnosis, such as brainstem infarct or posterior circulation stroke, without medullary-medial specification.}
\rubricitem{0}{Incorrect diagnosis unrelated to acute infarction of the medulla, or no diagnosis provided.}
\end{description}

\paragraph{Localisation reference.}
Laterality: bilateral. Organ/region: brainstem, specifically the medulla
oblongata. Specific substructure/segment: anteromedial/rostral medulla
oblongata.

\begin{description}
\setlength{\itemsep}{0.25em}
\rubricitem{3}{Bilateral anteromedial/rostral medulla oblongata.}
\rubricitem{2}{Identifies the medulla or brainstem but is incomplete on laterality or substructure.}
\rubricitem{1}{Broad posterior fossa, brainstem, or vertebrobasilar localisation without specifying the medulla.}
\rubricitem{0}{Incorrect or non-localising; outside the brainstem/posterior fossa, or no localisation.}
\end{description}

\subsubsection{Case 13789: Cerebral amyloid angiopathy-related inflammation}

\paragraph{Diagnosis reference.}
Cerebral amyloid angiopathy-related inflammation (CAA-ri).

\begin{description}
\setlength{\itemsep}{0.25em}
\rubricitem{3}{Specific diagnosis of cerebral amyloid angiopathy-related inflammation.}
\rubricitem{2}{Near-equivalent but less specific diagnosis, such as inflammatory CAA or CAA with vasogenic oedema, without clearly naming CAA-ri.}
\rubricitem{1}{Plausible alternative such as PRES, PML, seizure-related change, or cerebral amyloid angiopathy without the inflammatory/oedematous presentation.}
\rubricitem{0}{Incorrect diagnosis unrelated to the case pattern, or no diagnosis provided.}
\end{description}

\paragraph{Localisation reference.}
Laterality: bilateral, right greater than left. Organ/region: cerebral
hemispheric lobar white matter. Specific substructure/segment: subcortical
white matter at the cortical--subcortical interface of the posterior-predominant
parietal, occipital, and temporal lobes, most marked in the right temporal lobe.

\begin{description}
\setlength{\itemsep}{0.25em}
\rubricitem{3}{Bilateral posterior-predominant lobar cerebral involvement at the cortical--subcortical interface, most marked on the right.}
\rubricitem{2}{Bilateral cerebral hemispheric white-matter abnormality with posterior/lobar emphasis, but missing subcortical-interface detail or right-sided predominance.}
\rubricitem{1}{Broad brain or white-matter localisation without bilateral/lobar or subcortical-interface specificity.}
\rubricitem{0}{Incorrect or non-localising.}
\end{description}

\newpage
%

\lstdefinestyle{euroradprompt}{
  basicstyle=\ttfamily\scriptsize,
  breaklines=true,
  columns=fullflexible,
  keepspaces=true,
  showstringspaces=false,
  frame=single,
  framerule=0.25pt,
  xleftmargin=0.5em,
  xrightmargin=0.5em
}

\section{Prompt, Matching, and Evaluation Interface}
\label{app:prompt_matching_eval}

This appendix documents the prompts and interface settings used in the EuroRad diagnostic-workup benchmark.  Placeholders such as \texttt{\{case.clinical\_history\}}, \texttt{\{budget\}}, and \texttt{\{metadata\_block\}} indicate case-specific values filled before inference.  The evaluated agent never sees the hidden exam inventory, expert captions, key findings, final answers, or the reference differential diagnosis list, except in explicitly labeled oracle ablations.  Full case-level prompts, raw model outputs, normalized outputs, request-resolution logs, judge outputs, and summaries are stored in the released run logs.

\paragraph{Disclosure scope.}
We disclose the official sequential prompt templates, passive/oracle ablation prompts, LLM-judge prompt, deterministic rule-judge policy, structured output schemas, request-matching policy, and core runtime settings.  We do not enumerate every case-instantiated prompt in the PDF because these prompts repeat the same templates and differ mainly in clinical history, evidence metadata, request history, and attached images.

\begin{table*}[th]
\centering
\small
\setlength{\tabcolsep}{3.5pt}
\renewcommand{\arraystretch}{1.12}
\begin{tabular}{@{}p{0.18\linewidth}p{0.22\linewidth}p{0.30\linewidth}p{0.22\linewidth}@{}}
\toprule
Component & Model-visible input & Output / action & Key setting \\
\midrule
Official sequential agent & Clinical history; hidden-bundle count; prior-study hints when applicable; revealed images and minimal metadata after matched requests & Four-item differential at every turn; either one free-form imaging request or final stop answer with localization & Request budget \(B=6\); hidden inventory not listed; forced stop after budget \\
Request matcher & Free-form requested examination and hidden metadata pool & Matched evidence unit or unmatched/invalid request reason & Deterministic resolver, not an LLM; threshold \(5.0\), ambiguity margin \(0.75\) \\
LLM judge & Case-specific diagnosis/localization rubrics, reference differential set, final output, and trajectory diagnoses & 0--3 endpoint scores; E/A/U labels and 0--3 scores for trajectory diagnoses & Vertex Gemini 3 Flash in reported full runs; structured JSON output \\
Rule-based judge & Same case/model payload as the LLM judge & Deterministic 0--3 endpoint scores; deterministic E/A/U trajectory labels and scores & No LLM call \\
Passive ablations & Clinical history alone, all images at once, or passive random/gold evidence reveal & Stop-turn JSON after each passive state; no active requests & Route metrics structurally not applicable \\
Oracle-findings ablation & Official hidden-inventory request setting; matched evidence reveals images, metadata, and oracle findings & Same action schema as official sequential agent & Active request setting; route metrics applicable \\
\bottomrule
\end{tabular}
\caption{Prompted components and interface settings.  Case-specific fields are filled at runtime; image evidence is attached separately from the text prompt.}
\label{tab:prompt_interface_summary}
\end{table*}

\subsection{Official Sequential Agent Prompt}
\label{app:official_agent_prompt}

The official benchmark is an active evidence-acquisition task.  At the first turn, the agent receives only the clinical history, a count of hidden evidence bundles, and optionally a text notice that patient-provided prior imaging is available on request.  The hidden inventory list is never revealed.  At each subsequent turn, the agent receives the request history, the previous request-resolution result, and newly attached images if the previous request matched an evidence unit.

\paragraph{Official runtime settings.}
The official full runs use request budget \(B=6\), reveal unit \texttt{eurorad}, trajectory horizon \(T_{\max}=B+2=8\), and diagnostic threshold \(\tau=2/3\).  Candidate and judge maximum output tokens are both set to \(8192\).  Candidate-model provider temperature is not explicitly set and therefore uses provider defaults.  The LLM judge is run by Vertex Gemini 3 Flash with temperature 0.  The reported full runs use \texttt{judge\_modes=both}.

\textbf{Agent system prompt.}\par\noindent
\begin{lstlisting}[style=euroradprompt]
You are a rigorous radiology diagnostic agent.
You are participating in a benchmark with a hidden exam inventory.
Return STRICT JSON only. Do not use markdown, code fences, commentary, or extra keys.

Required JSON schema for every turn:
{
  "action": "request_exam" | "stop",
  "requested_examination": "free-form text; use an empty string when action=stop",
  "current_differential": [
    {"diagnosis": "string", "probability": 0.25},
    {"diagnosis": "string", "probability": 0.25},
    {"diagnosis": "string", "probability": 0.25},
    {"diagnosis": "string", "probability": 0.25}
  ],
  "final_location": {
    "laterality": "string",
    "region": "string",
    "substructure": "string"
  }
}

Rules:
- current_differential must contain exactly 4 UNIQUE diagnoses.
- Every probability must be in [0,1].
- The 4 probabilities must sum to 1.
- If action=request_exam, request exactly one EuroRad-style imaging evidence bundle in free-form natural language; final_location may contain empty strings.
- If action=stop, final_location must describe the lesion location and requested_examination must be empty.
- Each imaging request consumes one step, including unavailable, duplicate, or out-of-scope requests.
- Request imaging only when it is expected to help confirm, exclude, localize, stage, or characterize a relevant diagnosis.
- Request one bundle at a time. Do not combine multiple distinct body regions or modalities in one request.
- Include modality and anatomic region; include sequence/acquisition/contrast or prior-comparison timepoint when that distinction matters.
- The official requestable inventory includes current-presentation imaging and, when explicitly announced, patient-provided prior/comparison imaging. Records marked as future follow-up by numeric time_past < 0 are excluded from the official requestable/evaluable pool.
- Do not request imaging merely to exhaust the budget. Stop and submit a final answer when you have sufficient evidence.
- Do not mention candidate diagnosis lists or hidden inventory.
- Use only the clinical history, the revealed image evidence, and the revealed minimal exam metadata.
- Expert captions, key findings, and final answers are not provided to you.
\end{lstlisting}

\textbf{Initial turn user prompt template.}\par\noindent
\begin{lstlisting}[style=euroradprompt]
Clinical history:
{case.clinical_history or '[none]'}

Task context:
You are evaluating an unknown diagnostic imaging case. Imaging evidence is hidden until requested. Diagnose and localize the case by actively requesting available EuroRad-style figure/protocol-level imaging bundles when useful.

Official setting reminders:
- Hidden available exam bundles in this case: {available_bundle_count}.
- The hidden exam inventory list is NOT revealed.
- You may request at most {budget} imaging examinations in total.
- No candidate diagnosis list is given.
- Every request consumes one step, even if it is unavailable, duplicate, or out-of-scope.
- Available bundles may include current-presentation imaging and, when explicitly announced, patient-provided prior/comparison imaging; records marked as future follow-up by numeric time_past < 0 are excluded from the official requestable/evaluable pool.
{patient_provided_block}

This is the first decision turn. No imaging examination has been revealed yet.
Task: output your current four-item differential with probabilities, then either request one next examination bundle or stop if you already have enough evidence.
Respond in STRICT JSON only.
\end{lstlisting}

\paragraph{Patient-provided prior/comparison block.}
This block is inserted only when the case contains requestable prior/comparison imaging with numeric \texttt{time\_past > 0} that has not yet been revealed.  Future follow-up imaging with numeric \texttt{time\_past < 0} is excluded from the official requestable/evaluable pool.

\begin{lstlisting}[style=euroradprompt]
Patient-provided prior/comparison imaging available on request:
The patient brought the following prior imaging from before the current presentation. These images are not shown unless you request the relevant study.
- Available prior study {idx}: {modality}; {region}; {acquisition}; contrast: {contrast}; timepoint: {time_past}
If one of these studies would help, request it explicitly in natural language.
\end{lstlisting}

\paragraph{Update turn user prompt template.}\mbox{}\par\noindent
\begin{lstlisting}[style=euroradprompt]
Clinical history reminder:
{case.clinical_history or '[none]'}

Request budget used: {len(requests)} / {budget}
Remember: every request consumes one step, including unavailable, duplicate, or out-of-scope requests.

Request history so far:
{history_text}

{resolution_block}
{patient_provided_block}

Task:
Update your current four-item differential using all evidence seen so far, then choose the next action.
- If you still need evidence, set action=request_exam and request exactly one next exam bundle.
- If you are ready to conclude, set action=stop and provide final_location.

Respond in STRICT JSON only.
\end{lstlisting}

\textbf{Request-history and resolution text.}\par\noindent
\begin{lstlisting}[style=euroradprompt]
If a request matched:
- request #{req.request_index}: "{req.request_text}" -> MATCHED {req.matched_figure}

If a request was invalid:
- request #{req.request_index}: "{req.request_text}" -> INVALID ({req.invalid_reason})

Matched request resolution block:
Previous request result:
- MATCHED exam bundle: {exam.figure}
- Source figures: {', '.join(exam.source_figures or [exam.figure])}
- Newly revealed minimal metadata:
  - modality: {exam.modality or 'unspecified'}
  - acquisition: {exam.acquisition or 'unspecified'}
  - region: {exam.region or 'unspecified'}
  - contrast: {exam.contrast or 'unspecified'}
  - time_past: {exam.time_past or 'unspecified'}

New images for this matched exam are attached to this message.

Invalid request resolution block:
Previous request result:
- INVALID / UNMATCHED request
- Reason: {last_resolution.reason}
- No new exam bundle was revealed.
\end{lstlisting}

\textbf{Forced-stop user prompt template.}\par\noindent
\begin{lstlisting}[style=euroradprompt]
Clinical history reminder:
{case.clinical_history or '[none]'}

You have reached the maximum request budget ({budget} / {budget}).
You MUST stop now.

Request history:
{history_text}

{resolution_block}

Task:
Return a final stop-turn JSON. Set action=stop, provide your final four-item differential with probabilities, and include final_location.
Respond in STRICT JSON only.
\end{lstlisting}

\subsection{Request Matching Policy}
\label{app:request_matching_policy}

The request matcher is deterministic and does not use an LLM.  There is therefore no matcher prompt.  It maps each free-form requested examination to at most one hidden evidence unit using normalized text, modality, acquisition, view, region, contrast, figure identifiers, timepoint metadata, and rule-based ambiguity handling.

\textbf{Matcher function and thresholds.}\par\noindent
\begin{lstlisting}[style=euroradprompt]
resolve_request_to_exam(
    request_text,
    official_exam_pool,
    excluded_exam_pool,
    revealed_exam_ids,
    attempted_request_texts,
    match_threshold=5.0,
    ambiguity_margin=0.75,
)
\end{lstlisting}

\paragraph{Matcher outcomes.}
Unmatched/invalid outcomes include \texttt{empty\_request}, \texttt{duplicate\_request\_text}, \texttt{no\_official\_exam\_pool}, \texttt{already\_revealed\_exam\_requested}, \texttt{unavailable\_exam\_requested}, and \texttt{no\_match\_above\_threshold}.  Matched outcomes include \texttt{matched}, \texttt{matched\_timepoint\_tiebreak}, and several best-effort ambiguous-match reasons.  Ambiguous or broad requests are not automatically penalized as invalid: if at least one eligible unrevealed official evidence unit scores above threshold, the resolver reveals the best-scoring eligible candidate and logs \texttt{ambiguity\_resolved}, \texttt{candidate\_scores}, and \texttt{resolution\_reason}.

\paragraph{Follow-up and prior-imaging policy.}
Numeric \texttt{time\_past < 0} denotes future follow-up imaging and is excluded from the requestable/evaluable official pool.  Numeric \texttt{time\_past > 0} denotes patient-provided prior imaging and remains requestable; the agent is informed that such prior imaging is available on request.  Null or nonnumeric \texttt{time\_past} does not by itself expose or exclude an exam.

\subsection{LLM Judge Prompt}
\label{app:llm_judge_prompt}

The LLM judge is a single-call text-only judge that receives the case-specific rubrics, reference differential set, final model output, and all unique diagnosis strings appearing in the model trajectory.  In reported full runs, the LLM judge is Vertex Gemini 3 Flash.  A deterministic rule judge is also run for audit/agreement analysis, but the reported main table uses the LLM-judge mode.

\textbf{Judge system prompt.}\par\noindent
\begin{lstlisting}[style=euroradprompt]
You are a strict clinical benchmark judge for multimodal differential diagnosis.
Return STRICT JSON only. Do not use markdown or extra keys.

You will score:
1) final diagnosis quality using the provided case-specific diagnosis rubric;
2) final localization quality using the provided case-specific localization rubric;
3) final four-item differential-list quality using the reference differential set and the global rubric below;
4) exact/acceptable/unmatched labels and 0-3 diagnosis-rubric scores for every diagnosis string in the trajectory.

Global rubric for final differential-list quality (0-3):
- 0: The list is mostly off-target, fails to include the final diagnosis or close equivalent, and has little overlap with the reference differential set.
- 1: The list contains one or more accepted-but-not-gold items, but coverage/ranking is weak and the list does not function as a strong clinical differential.
- 2: The list includes the final diagnosis or a near-gold diagnosis, but coverage or ranking is incomplete/suboptimal.
- 3: The correct diagnosis is prominent, and the remaining items are largely aligned with reference_differential_options or explicit case-rubric examples.

Trajectory labels and scores:
- E (exact): reserved exclusively for the gold final diagnosis concept, including close lexical variants or true near-synonyms of the gold diagnosis. Do NOT label a non-gold reference differential option as E.
- A (acceptable): not exact, but accepted for this case because it matches reference_differential_options, is a close synonym of such an option, or would receive score 1 or 2 under the case-specific diagnosis rubric.
- U (unmatched): not exact and not accepted by the reference differential set or diagnosis rubric. Do not mark a diagnosis A merely because it is generically clinically plausible.
- trajectory_scores.score must use the same case-specific 0-3 diagnosis rubric as the final diagnosis score. These scores are logged for rubric-based trajectory diagnostics and time-to analyses. Confidence-weighted trajectory metrics use the E/A/U labels together with the model's reported probabilities.

Be conservative, concise, and consistent.
\end{lstlisting}

\textbf{Judge user prompt template.}\par\noindent
\begin{lstlisting}[style=euroradprompt]
Score the model output for this case.

CASE PAYLOAD
{json_dumps(case_payload)}

MODEL PAYLOAD
{json_dumps(model_payload)}

Instructions:
- Use the provided case-specific diagnosis rubric to assign diagnosis.score in {0,1,2,3}.
- Use the provided case-specific localization rubric to assign localization.score in {0,1,2,3}.
- Use the global differential-list rubric from the system prompt to assign differential_list.score in {0,1,2,3}.
- For the differential-list score, treat reference_differential_options as the reference differential set, supplemented only by explicit examples in the case-specific diagnosis rubric. The model list is final_differential.
- Penalize non-reference diagnoses even if they are generically plausible, unless the case-specific diagnosis rubric would clearly award them score 1 or 2.
- For each unique diagnosis string in trajectory_unique_diagnoses, assign exactly one label: E, A, or U. E is only for the gold final diagnosis concept.
- For each unique diagnosis string in trajectory_unique_diagnoses, also assign trajectory_scores.score in {0,1,2,3} using the provided case-specific diagnosis rubric.
- Provide a brief reason (<=30 words) for each final score and each trajectory score.
- Every diagnosis from trajectory_unique_diagnoses must appear exactly once in trajectory_labels and exactly once in trajectory_scores.

Return STRICT JSON only with this schema:
{
  "final_scores": {
    "diagnosis": {"score": 0, "reason": "string"},
    "localization": {"score": 0, "reason": "string"},
    "differential_list": {"score": 0, "reason": "string"}
  },
  "trajectory_labels": [
    {"diagnosis": "string", "label": "E|A|U", "reason": "string"}
  ],
  "trajectory_scores": [
    {"diagnosis": "string", "score": 0, "reason": "string"}
  ]
}
\end{lstlisting}

\subsection{Rule-based Judge}
\label{app:rule_based_judge}

The deterministic judge is implemented in code as \texttt{RuleScorer}.  It is not prompted and does not call an LLM.  It receives the same normalized case/model payload as the LLM judge: the final top-1 diagnosis, the final four-item differential list, the structured final localization, and the set of unique diagnosis strings that appeared anywhere in the model trajectory.  It returns the same output schema as the LLM judge, including final 0--3 scores, trajectory E/A/U labels, and trajectory 0--3 diagnosis-rubric scores used for rubric-based trajectory diagnostics and time-to analyses.  In \texttt{judge\_modes=both}, both judges are run; the default reported mode is the LLM judge when available, while the rule judge is logged for reproducibility and agreement diagnostics.  If LLM judging is disabled or fails while rule judging is enabled, the rule result is used as the available judge result.

\paragraph{Diagnosis bucket construction.}
For each case, the rule judge first builds three diagnosis-matching buckets from the case-specific diagnosis rubric and reference differential:
\begin{itemize}
    \item \textbf{Score-3 / exact-gold bucket}: the diagnosis rubric \texttt{reference\_answer}, falling back to \texttt{final\_diagnosis}, plus explicitly extracted score-3 terms from the rubric text.
    \item \textbf{Score-2 / near-gold bucket}: explicitly extracted score-2 rubric terms.  If no such terms are available, relaxed aliases of the gold diagnosis are used.
    \item \textbf{Score-1 / acceptable-differential bucket}: non-gold entries from \texttt{reference\_ddx\_options}, plus explicitly extracted score-1 rubric terms.
\end{itemize}
Rubric terms are extracted conservatively from quoted phrases and marker phrases such as ``acceptable terms include'', ``near-gold includes'', ``such as'', ``e.g.'', and ``for example''.  The rule judge does not invent new clinical alternatives beyond the case rubric, the gold answer, and the reference differential set.

\paragraph{Normalization and alias matching.}
Before matching, diagnosis strings are normalized by applying spelling/canonicalization maps, removing common filler phrases, stripping punctuation and brackets, normalizing hyphens and whitespace, and extracting only the primary diagnosis concept from a list-like answer.  Negated or empty concepts are treated as invalid.  For each reference term, aliases include the normalized surface form, parenthesis-stripped form, acronym/initialism variants when present, hyphen/space-collapsed variants, and relaxed variants with low-specificity qualifier tokens removed.  Near-gold matching also allows exact equality after removing qualifier tokens, subset/superset overlap among informative gold tokens, and disease-family matches with sufficient lexical similarity.  Reference-option matching uses exact alias matches first, then high lexical similarity, informative token overlap, disease-family overlap, and anatomy-equivalence overlap.

\paragraph{Top-1 diagnosis score.}
The rule-based top-1 diagnosis score is assigned as follows:
\begin{itemize}
    \item \textbf{3}: the model's primary diagnosis exactly matches the gold diagnosis concept or a very close lexical alias in the score-3 bucket.
    \item \textbf{2}: the diagnosis is near-gold but missing a qualifier or specificity, or matches the score-2 / relaxed-gold bucket.
    \item \textbf{1}: the diagnosis is not gold, but matches a reference differential option or close equivalent in the score-1 bucket.
    \item \textbf{0}: the diagnosis is empty, negated, invalid, or off-target.
\end{itemize}

\paragraph{Trajectory labels and trajectory scores.}
Every unique diagnosis string produced during the trajectory is scored using the same top-1 diagnosis rule above.  The E/A/U label is derived from the same buckets:
\begin{itemize}
    \item \textbf{E}: exact gold diagnosis concept or very close gold lexical variant.
    \item \textbf{A}: near-gold diagnosis or accepted reference-differential option.
    \item \textbf{U}: unmatched, invalid, or off-target diagnosis.
\end{itemize}
This design keeps the logged trajectory diagnosis scores aligned with the endpoint diagnosis rubric rather than using a separate trajectory-specific rule; the confidence-weighted trajectory metrics use the E/A/U labels together with model probabilities.

\paragraph{Localization score.}
The rule judge scores \texttt{final\_location} against the structured localization rubric reference answer.  It separately normalizes laterality, organ/region, and substructure.  Laterality aliases are mapped to canonical modes such as left, right, bilateral, midline, unilateral, or none.  Region and substructure are tokenized after removing generic region stopwords and expanded with predefined anatomy equivalences, for example brainstem/medulla/pons/midbrain, posterior fossa/cerebellum/brainstem, white matter/subcortical, cortex/cortical, ventricle/fourth ventricle, and spinal cord/cord/myelon.  The score is:
\begin{itemize}
    \item \textbf{3}: correct region, compatible laterality, and full specific-substructure overlap.
    \item \textbf{2}: correct region with compatible or partially compatible laterality and partial localization specificity, without a specific substructure conflict.
    \item \textbf{1}: broad region or anatomy overlap, but incomplete localization.
    \item \textbf{0}: wrong or absent localization.
\end{itemize}

\paragraph{Final differential-list score.}
The final differential-list score uses only the first four diagnoses in the model's final list.  Each item is labeled by the same E/A/U diagnosis matcher.  The rule judge records the rank of the first exact-gold diagnosis, the rank of the first near-gold diagnosis, and the number of unique acceptable reference-differential matches.  The score is:
\begin{itemize}
    \item \textbf{3}: the exact gold diagnosis is ranked first and at least three unique list items align with the reference differential set.
    \item \textbf{2}: the exact gold diagnosis appears anywhere, or a near-gold diagnosis appears in rank 1--2, and at least two unique list items are acceptable.
    \item \textbf{1}: at least one list item matches an acceptable reference differential option.
    \item \textbf{0}: the list is mostly off-target.
\end{itemize}

\paragraph{Dual-mode agreement logging.}
When both LLM and rule judges are enabled, the runner stores both outputs under separate \texttt{by\_mode} entries and computes agreement diagnostics.  These include exact score agreement for diagnosis, localization, and differential-list endpoint scores; trajectory-label agreement; and mean absolute error between LLM and rule trajectory diagnosis scores.  These diagnostics are logged but do not change the reported LLM-judge result when the LLM judge succeeds.

\subsection{Ablation Prompt Templates}
\label{app:ablation_prompts}

The ablations use the same case schema and judge, but change how evidence is exposed.  Passive ablations disallow active imaging requests, so route/request metrics such as essential recall and order concordance are structurally not applicable.  The oracle-findings ablation remains an active request-based setting; matched evidence reveals images, minimal metadata, and oracle findings for that matched evidence unit.

\textbf{Ablation settings.}
\begin{itemize}
    \item \textbf{History-only}: clinical history only, no images.
    \item \textbf{All-images-at-once}: all requestable evidence units are attached in one call.
    \item \textbf{Random-order reveal}: evidence units are passively revealed one at a time in seeded random order.
    \item \textbf{Gold-order reveal}: evidence units are passively revealed one at a time by preferred/gold order.
    \item \textbf{Oracle findings}: official hidden-inventory request setting, but matched evidence reveals oracle findings.
\end{itemize}

\textbf{Passive ablation system prompt.}\par\noindent
\begin{lstlisting}[style=euroradprompt]
You are a rigorous radiology diagnostic agent.
Return STRICT JSON only. Do not use markdown, code fences, commentary, or extra keys.

Required JSON schema:
{
  "action": "stop",
  "requested_examination": "",
  "current_differential": [
    {"diagnosis": "string", "probability": 0.25},
    {"diagnosis": "string", "probability": 0.25},
    {"diagnosis": "string", "probability": 0.25},
    {"diagnosis": "string", "probability": 0.25}
  ],
  "final_location": {
    "laterality": "string",
    "region": "string",
    "substructure": "string"
  }
}

Rules:
- action must be "stop". These ablations do not allow active imaging requests.
- current_differential must contain exactly 4 UNIQUE diagnoses.
- Every probability must be in [0,1], and the 4 probabilities must sum to 1.
- Provide the best current diagnosis and lesion localization using only the evidence given so far.
- No candidate diagnosis list is given.
\end{lstlisting}

\textbf{History-only prompt template.}\par\noindent
\begin{lstlisting}[style=euroradprompt]
Setting: History-only ablation.

Clinical history:
{case.clinical_history or '[none]'}

No imaging is attached and no imaging may be requested in this ablation. Use only the clinical history.

Task:
Return one final stop-turn JSON with your four-item differential diagnosis, probabilities, and final_location.
\end{lstlisting}

\textbf{All-images-at-once prompt template.}\par\noindent
\begin{lstlisting}[style=euroradprompt]
Setting: All-images-at-once ablation.

Clinical history:
{case.clinical_history or '[none]'}

All requestable EuroRad-style figure/protocol evidence units for this case are attached in one call. Expert captions, oracle imaging findings, and final answers are not provided. The list below gives only minimal evidence metadata and the 1-indexed order of attached images.

Attached evidence metadata:
{metadata_block}

Task:
Use the clinical history and all attached images to return one final stop-turn JSON with your four-item differential diagnosis, probabilities, and final_location.
\end{lstlisting}

\textbf{Passive random/gold reveal templates.}\par\noindent
\begin{lstlisting}[style=euroradprompt]
Initial passive turn:

Setting: {setting} ablation.

Clinical history:
{case.clinical_history or '[none]'}

This ablation passively reveals EuroRad-style figure/protocol evidence units. You cannot request imaging. The reveal order is controlled by the benchmark. Total planned evidence reveals: {n_reveals}.

This is the history-only baseline turn before the first evidence unit is revealed.

Task:
Return one stop-turn JSON with your current four-item differential diagnosis, probabilities, and final_location.

Reveal turn:

Setting: {setting} ablation.

Clinical history reminder:
{case.clinical_history or '[none]'}

The benchmark is passively revealing evidence units in {order_description}. You cannot request imaging.

Reveal {step_index} / {total_steps}:
{metadata_line}

New images for this evidence unit are attached to this message. Expert captions, oracle imaging findings, and final answers are not provided.

Task:
Update your current four-item differential diagnosis, probabilities, and final_location using all evidence seen so far. Return STRICT JSON with action=stop.
\end{lstlisting}

\textbf{Evidence metadata line template.}\par\noindent
\begin{lstlisting}[style=euroradprompt]
- images [{index_text}] | evidence_id={exam.exam_id} | figure={exam.figure} | source_figures={source_figures} | modality={modality} | acquisition={acquisition} | view={view} | region={region} | contrast={contrast} | time_past={time_past}
\end{lstlisting}

\textbf{Oracle-findings system prompt.}

\begin{lstlisting}[style=euroradprompt]
You are a rigorous radiology diagnostic agent.
You are participating in a EuroRad-style hidden evidence benchmark.
Return STRICT JSON only. Do not use markdown, code fences, commentary, or extra keys.

Required JSON schema for every turn:
{
  "action": "request_exam" | "stop",
  "requested_examination": "free-form text; use an empty string when action=stop",
  "current_differential": [
    {"diagnosis": "string", "probability": 0.25},
    {"diagnosis": "string", "probability": 0.25},
    {"diagnosis": "string", "probability": 0.25},
    {"diagnosis": "string", "probability": 0.25}
  ],
  "final_location": {
    "laterality": "string",
    "region": "string",
    "substructure": "string"
  }
}

Rules:
- current_differential must contain exactly 4 UNIQUE diagnoses.
- Every probability must be in [0,1], and the 4 probabilities must sum to 1.
- If action=request_exam, request exactly one next EuroRad-style figure/protocol evidence unit.
- Use modality, anatomic region, and when relevant sequence/acquisition/contrast/timepoint (e.g. T2-weighted MRI spine, DWI/ADC MRI brain, CTA, post-contrast T1, non-contrast CT).
- If action=stop, final_location must describe the lesion location and requested_examination must be empty.
- Each request consumes one step, including unavailable, duplicate, or out-of-scope requests.
- Ambiguous/broad requests may be resolved best-effort by the matcher and are not treated as agent errors when an eligible candidate exists.
- In this oracle-findings ablation, matched evidence reveals images, minimal metadata, and oracle imaging findings for the matched evidence unit.
- No candidate diagnosis list is given.
\end{lstlisting}

\textbf{Oracle-findings user prompt templates.}\par\noindent
\begin{lstlisting}[style=euroradprompt]
Initial:

Setting: Oracle-findings ablation.

Clinical history:
{case.clinical_history or '[none]'}

Official setting reminders:
- The hidden EuroRad-style evidence inventory list is NOT revealed.
- Hidden available evidence units in this case: {len(case.official_exam_pool)}.
- You may request at most {budget} evidence units in total.
- Every request consumes one step, even if it is unavailable, duplicate, ambiguous/best-effort resolved, or out-of-scope.
- When a request matches an evidence unit, images, minimal metadata, and oracle imaging findings for that matched unit will be revealed.

This is the first decision turn. No imaging evidence has been revealed yet.
Task: output your current four-item differential with probabilities, then either request the next evidence unit or stop if you already have enough evidence.
Respond in STRICT JSON only.

Update:

Clinical history reminder:
{case.clinical_history or '[none]'}

Request budget used: {len(requests)} / {budget}
Remember: every request consumes one step, including unavailable, duplicate, or out-of-scope requests.

Request history so far:
{history_text}

{resolution_block}
Task:
Update your current four-item differential using all evidence seen so far, then choose the next action.
- If you still need evidence, set action=request_exam and request exactly one next EuroRad-style evidence unit.
- If you are ready to conclude, set action=stop and provide final_location.

Respond in STRICT JSON only.
\end{lstlisting}

\textbf{Oracle matched-resolution block.}

\begin{lstlisting}[style=euroradprompt]
Previous request result:
- MATCHED evidence unit: {exam.figure}
- Source figures: {', '.join(exam.source_figures or [exam.figure])}
- Minimal metadata:
  - modality: {exam.modality or 'unspecified'}
  - acquisition: {exam.acquisition or 'unspecified'}
  - region: {exam.region or 'unspecified'}
  - contrast: {exam.contrast or 'unspecified'}
  - time_past: {exam.time_past or 'unspecified'}

Oracle imaging findings for this matched evidence unit:
{format_oracle_findings(case, exam)}

New images for this matched evidence unit are attached to this message.
\end{lstlisting}

\subsection{Metric and Undefined-Value Settings}
\label{app:metric_settings}

Endpoint scores are normalized from 0--3 judge rubric scores: \(S_{\mathrm{dx}}\), \(S_{\mathrm{loc}}\), and \(S_{\mathrm{ddx}}\) equal the corresponding judge score divided by 3.  Essential recall, optional burden, unmatched-request rate, and order concordance are computed from matched and unmatched request events against the hidden official evidence pool.  Passive ablation settings do not contain model requests, so route/request metrics are displayed as structurally undefined.

Trajectory confidence alignment follows the main-text definition.  At each turn, the judge labels diagnoses as exact-match, acceptable differential, or unmatched; \(S_{\mathrm{conf}}^{(t)}\) rewards probability mass on exact or acceptable diagnoses and penalizes probability mass on unmatched diagnoses.  \(S_{\mathrm{traj}}\) is the trajectory average of \(S_{\mathrm{conf}}^{(t)}\) over evaluated turns.  Time-to-diagnostic-guess uses threshold \(\tau=2/3\).  If a time-to event is never reached, the logged time-to value is \(T_{\max}+1\); efficiency analyses also report the reached-case proportion for clinically acceptable diagnosis.

\clearpage
\section{Benchmark Card and Responsible Release}
The benchmark release includes a benchmark card alongside the JSON data, scoring code, prompts, result manifests, Croissant metadata, and attribution files. The card is intended to make provenance, annotation scope, intended use, non-use, risks, access, and maintenance explicit for the public release.

\subsection{Benchmark card fields}
\begin{table*}[!htbp]
    \centering
    \scriptsize
    \setlength{\tabcolsep}{3pt}
    \renewcommand{\arraystretch}{1.08}
    \begin{tabular}{p{0.19\linewidth}p{0.45\linewidth}p{0.28\linewidth}}
        \toprule
        Field group & Fields to report & Current entry / source \\
        \midrule
        Identity &
        Benchmark name, version, release date, clinical domain, source collection, number of cases, evidence units, diagnostic steps, and images &
        \bench{} v1.0.0; 2026-05-06; neuroradiology; EuroRad-derived; 211 annotated candidates; 191 retained release/evaluation cases; 811 imaging-examination metadata records; 785 route-evaluable/requestable evidence units; 789 diagnostic-step annotations; 1,609 images \\

        Access and code &
        Dataset URL, code URL, hosted artifacts, release manifests, and scoring-code repository &
        Dataset: \url{https://huggingface.co/datasets/Anonym001/DDx-TRACE}; Code: \url{https://github.com/DDx-Trace/DDx-TRACE} \\

        Provenance &
        Source URLs, source case titles, source publication dates, source image paths, reconstruction or download instructions &
        Stored in release JSON, \texttt{images.csv}, \texttt{ATTRIBUTION.tsv}, and Croissant metadata \\

        Annotation team &
        Annotator names or anonymized roles, clinical background, assignment policy, review independence, conflict-resolution policy &
        Three primary physician annotators performed independent case reviews, and two
additional board-certified senior physicians contributed to adjudication and meta-review. \\

        Release filtering &
        Candidate cases, exclusion count, retained cases, filtering status, and case-level exclusion rationale &
        211 candidate cases were annotated; 20 unsuitable cases were excluded; 191 cases are retained in the release/evaluation set \\

        Case labels &
        \texttt{difficulty}, \texttt{rarity}, \texttt{discard}, free-text case comments, and demography comments &
        Difficulty: 115 normal, 73 hard, 3 extreme hard; rarity: 17 common, 146 rare, 28 extreme rare \\

        Diagnostic-step labels &
        Step text, source figure, reasoning note, importance label, reviewer comments, final adjudicated value &
        789 records; labels include \texttt{essential}, \texttt{optional}, and \texttt{unnecessary} where applicable \\

        Exam labels &
        Figure ID, caption, modality, preferred order, metadata fields, reviewer metadata feedback, final adjudicated metadata &
        811 metadata records; 785 are route-evaluable/requestable after excluding 26 numeric-future-follow-up records; fields include modality, acquisition, view, region, \texttt{time\_past}, contrast, and order \\

        Image labels and attribution &
        Image/subfigure ID, image path, caption, source image URL, source case URL, and attribution metadata &
        1,609 image/subfigure records; attribution stored in \texttt{images.csv} and \texttt{ATTRIBUTION.tsv} \\

        Rubric labels &
        Diagnosis reference answer, diagnosis 0--3 rubric, localization reference components, localization 0--3 rubric, rubric feedback and confirmed edits &
        Stored in \texttt{rubric\_0\_to\_3} fields; exported schema retains \texttt{Localisation} as the key for localization rubrics \\
        \bottomrule
    \end{tabular}
    \caption{\textbf{Benchmark card fields for responsible release, part 1.} Core \bench{} v1.0.0 identity, access, provenance, annotation, and label fields, completed from the provided release JSON and Croissant metadata.}
    \label{tab:benchmark_card_fields}
\end{table*}

\begin{table*}[!htbp]
    \centering
    \scriptsize
    \setlength{\tabcolsep}{3pt}
    \renewcommand{\arraystretch}{1.08}
    \begin{tabular}{p{0.19\linewidth}p{0.45\linewidth}p{0.28\linewidth}}
        \toprule
        Field group & Fields to report & Current entry / source \\
        \midrule
        Evaluation protocol &
        Agent input, hidden information, request budget, reveal unit, output schema, judge model, scoring modes, metric definitions &
        Reported per run in prompts, configs, result manifests, and metric notes \\

        Croissant and RAI metadata &
        Croissant core metadata, Responsible AI metadata, validation status, reviewer-accessible metadata URL &
        \texttt{croissant.json} provided with dataset URL, distribution files, checksums, license, and RAI fields \\

        Intended use &
        Research evaluation of multimodal diagnostic workup trajectories; model comparison; error analysis; process-aware benchmark development &
        Research benchmark only; not clinical decision support \\

        Non-use &
        Direct patient care, triage, treatment decisions, credentialing clinicians, or claims of clinical safety without separate validation &
        Explicitly prohibited in release card and Croissant RAI metadata \\

        Risks and limitations &
        Published-case leakage, shortcut use of textual histories, incomplete clinical context, publication bias, image subset rather than full studies, neuroradiology-only scope &
        Documented in Croissant RAI metadata and benchmark card \\

        License and maintenance &
        License terms, changelog, versioning, contact, issue-reporting mechanism, deprecation policy &
        Dataset license: \url{https://creativecommons.org/licenses/by-nc-sa/4.0/}; code repository and release artifacts are versioned with checksums and a changelog \\
        \bottomrule
    \end{tabular}
    \caption{\textbf{Benchmark card fields for responsible release, part 2.} Evaluation, Croissant/RAI, intended-use, non-use, risk, licensing, and maintenance fields for \bench{} v1.0.0.}
    \label{tab:benchmark_card_fields_release}
\end{table*}

\subsection{Annotation and meta-review fields}

\begin{table*}[!htbp]
    \centering
    \small
    \begin{tabular}{p{0.20\linewidth}p{0.34\linewidth}p{0.38\linewidth}}
        \toprule
        Field family & Stored fields & Documentation requirement \\
        \midrule
        Reviewer provenance & \texttt{doctor\_annotations}, source file, dataset ID, export timestamp, anonymized annotator ID & Identify which two anonymized physician reviewers (Reviewer A/B/C) reviewed each case; reviewer identities should remain anonymized in the public/submission release \\
        Case-level review & \texttt{difficulty}, \texttt{rarity}, \texttt{discard}, \texttt{comment}, \texttt{red\_flag\_exam}, demography note & Define label options and how conflicts are resolved in the final metadata \\
        Step-level review & \texttt{steps\_essential}, step source, step reason, final step text & Define essential vs.\ optional vs.\ unnecessary and how labels affect route metrics \\
        Exam-order review & \texttt{imaging\_preferred\_order}, final \texttt{preferred\_order} & State whether order is a strict ranking or partial order with ties; document aggregation/adjudication rule \\
        Metadata review & \texttt{imaging\_meta\_info\_ok}, \texttt{imaging\_meta\_info\_comment}, \texttt{meta\_info\_updates} & Document modality/acquisition/view/region/\texttt{time\_past}/contrast correction workflow \\
        Rubric review & \texttt{localisation\_rubric\_ok}, \texttt{diagnosis\_rubric\_ok}, rubric comments, \texttt{rubric\_updates} & Document how diagnosis and localization rubrics were corrected and confirmed \\
        Meta-review status & \texttt{auto\_resolutions}, \texttt{green\_light}, \texttt{last\_updated}, confirmed update flags & Report green-light criteria, unresolved issues, and final meta-review completion rate \\
        Release filtering & \texttt{Discard}, missing-image checks, follow-up/prior exam handling, leakage audit status & State which cases/exams are included, excluded, or retained only as provenance \\
        \bottomrule
    \end{tabular}
    \caption{\textbf{Annotation fields captured by the review interfaces.} Field names follow the annotator JSON and meta-review export schema.}
    \label{tab:annotation_release_fields}
\end{table*}
